**A Systematic Evaluation of the COTQ Provincial Land Cover Product: Structural Consistency, Spectral Separability, and Relative Positioning Against ESA, ESRI, and Google Products**

**Étienne Clabaut [1]*, Samuel Foucher [2], Yacine Bouroubi [3]**

Département de Géomatique Appliquée, Université de Sherbrooke, Sherbrooke, QC J1K 2R1, Canada; etienne.clabaut@usherbrooke.ca
Département de Géomatique Appliquée, Université de Sherbrooke, Sherbrooke, QC J1K 2R1, Canada; samuel.foucher@usherbrooke.ca
Département de Géomatique Appliquée, Université de Sherbrooke, Sherbrooke, QC J1K 2R1, Canada; yacine.bouroubi@usherbrooke.ca

Correspondence: etienne.clabaut@usherbrooke.ca

## Abstract

Global 10-m land-cover products derived from Sentinel-2 provide valuable large-scale information, but their standardized design may limit their suitability for regional monitoring objectives. In Québec, these limitations motivated the development of the Cartographie de l'Occupation du Territoire Québécois (COTQ), a provincial 10-m product designed for annual monitoring of land occupation and soil artificialisation. This study evaluates COTQ relative to ESA WorldCover, ESRI Land Cover, and Google Dynamic World using complementary structural, spatial, spectral, and expert-based analyses across eight study areas spanning Québec's main bioclimatic domains. The results consistently identify COTQ and WorldCover as the strongest-performing pair, while ESRI Land Cover and Dynamic World exhibit greater spatial generalization and weaker spectral coherence. A targeted photo-interpretation of substantive COTQ–WorldCover disagreement areas further shows that COTQ is more frequently supported for most evaluated classes, whereas WorldCover performs better in some specific ecological contexts. These results do not provide a province-wide overall accuracy estimate, but they provide converging evidence that COTQ offers a relative performance and operational advantage for the Québec conditions, harmonized classes, and monitoring objectives evaluated in this study.

The COTQ map and all reference products are publicly available for visual inspection at: https://ee-dga-couverture.projects.earthengine.app/view/cotq

**Évaluation systématique du produit provincial de couverture du sol COTQ : cohérence structurelle, séparabilité spectrale et positionnement relatif par rapport aux produits ESA, ESRI et Google**

## Résumé

Les produits globaux de couverture du sol à 10 m dérivés de Sentinel-2 fournissent une information précieuse à grande échelle, mais leur conception standardisée peut limiter leur adéquation à certains objectifs de suivi régional. Au Québec, ces limites ont motivé le développement de la Cartographie de l'Occupation du Territoire Québécois (COTQ), un produit provincial à 10 m conçu pour le suivi annuel de l'occupation du territoire et de l'artificialisation des sols. Cette étude évalue la COTQ par rapport à ESA WorldCover, ESRI Land Cover et Google Dynamic World à l'aide d'analyses structurelles, spatiales, spectrales et expertes complémentaires menées sur huit zones d'étude couvrant les principaux domaines bioclimatiques du Québec. Les résultats identifient de manière cohérente COTQ et WorldCover comme les deux produits les plus performants, tandis qu'ESRI Land Cover et Dynamic World présentent une généralisation spatiale plus importante et une cohérence spectrale plus faible. Une photo-interprétation ciblée des zones de désaccord substantiel entre COTQ et WorldCover montre également que COTQ est plus fréquemment soutenue pour la majorité des classes évaluées, alors que WorldCover se distingue dans certains contextes écologiques particuliers. Ces résultats ne constituent pas une estimation de l'exactitude globale à l'échelle du Québec, mais fournissent des éléments convergents en faveur d'un avantage relatif et opérationnel de COTQ dans les conditions environnementales, les classes harmonisées et les objectifs de suivi évalués dans cette étude.

La carte COTQ et l'ensemble des produits de référence sont accessibles publiquement pour inspection visuelle à l'adresse suivante : https://ee-dga-couverture.projects.earthengine.app/view/cotq

**Key Policy Highlights**

- The COTQ provincial land-cover product provides a more detailed representation of artificial land transformation and appears particularly well suited to Québec's operational monitoring needs.
- Multi-criteria evaluation combining structural and spectral analyses reveals that ESA WorldCover is the closest global analogue to COTQ, while still underestimating heterogeneous urban and industrial surfaces.
- The proposed evaluation framework offers a transferable methodology for regional land-cover validation in the absence of authoritative ground truth.
- The results demonstrate the value of developing a region-specific land-cover product such as COTQ to address ecological and operational requirements that are not fully captured by globally standardized datasets.

## 1. Introduction

"Essentially, all models are wrong, but some are useful"(Box and Draper, 1987). Land-cover maps are no exception: they are necessarily simplified representations of complex and continuously changing landscapes, constrained by spatial resolution, thematic definitions, observation conditions, and intended applications. Their value therefore lies not in providing a perfect representation of reality, but in providing a sufficiently reliable and useful representation for a given purpose. Land use and land cover (LULC) information is essential for environmental monitoring, natural resource management, biodiversity assessment, and climate-adaptation planning. In recent years, global initiatives based on Sentinel-2 imagery such as ESA WorldCover, Google Dynamic World, ESRI Land Cover, and other 10-m products have substantially improved the availability of high-resolution land-cover data worldwide (Benhammou et al., 2022; Brown et al., 2022; Karra et al., 2021; Venter et al., 2022). Earlier large-scale Sentinel-2 mapping initiatives, such as the Sentinel-2 Global Land Cover (S2GLC) project, also demonstrated the feasibility of automated multi-temporal land-cover production at 10-m resolution over continental extents (Malinowski et al., 2020). These efforts reflect a broader shift towards continental- and global-scale mapping using standardized Earth observation data, a trend emphasized by recent reviews and community roadmaps calling for sustained medium-resolution products derived from the Landsat–Sentinel constellation (Radeloff et al., 2024).

Although all of these global products rely on Sentinel-2 data, they differ substantially in their underlying modelling strategies. ESA WorldCover adopts a hybrid system combining machine-learning classifiers, temporal statistics, and rule-based refinements (Zanaga et al., 2022), whereas Dynamic World and ESRI Land Cover rely on end-to-end deep convolutional networks trained on massive reference datasets (Brown et al., 2022; Karra et al., 2021). These heterogeneous pipelines lead to differences in spatial smoothness, class definitions, and sensitivity to local ecological variation, which reinforces the need for regional evaluations.

These global products have been subjected to extensive producer-led and independent validation campaigns and therefore constitute established benchmarks for 10-m land-cover mapping. ESA WorldCover 2021 v200, for example, achieved a global overall accuracy of 76.7 ± 0.5% in its official independent validation (Tsendbazar, 2022). A more recent independent comparative assessment of WorldCover, Dynamic World, and ESRI Land Cover reported a global overall accuracy of 83.8 ± 0.4% for WorldCover under the authors' recommended validation approach, compared with 74.1 ± 0.6% for Dynamic World and 73.4 ± 0.7% for ESRI Land Cover, and identified WorldCover as the highest-performing of the three products across all continents (Xu et al., 2024). Previous comparative analyses have also reported generally strong performance of these global products in North America (Venter et al., 2022). Although such published accuracy estimates cannot be transferred directly to Québec because validation designs, reference datasets, and class definitions differ, they establish that the products used here, and WorldCover in particular, provide meaningful externally validated benchmarks against which the performance of COTQ can be assessed.

Despite their global value, existing 10-m LULC products show limitations for regions with complex ecological gradients, heterogeneous surface materials, and strong phenological variability. Comparative assessments have revealed substantial discrepancies in boreal, subarctic, and temperate forest environments (Xu et al., 2024; Zhao et al., 2023). Persistent issues include confusion between bare rock, sparse vegetation, and cryptogamic substrates on the Precambrian Shield; misclassification of wetlands and peatland ecotones; difficulties in resolving small anthropogenic features; and temporal instability linked to seasonal snow, shadows, and cloud contamination. Moreover, global products are designed primarily for worldwide uniformity rather than high fidelity in any particular landscape context, which prevents them from reflecting locally defined class schemes, thematic priorities, or operational management needs.

These limitations also apply in the province of Québec, where the Ministère des Ressources naturelles et des Forêts (MRNF) requires a province-wide 10 m land-cover map updated annually to monitor land occupation in general and soil artificialisation in particular. Urban expansion, infrastructure development, and land conversion often occur in small, spatially fragmented patterns that global products struggle to capture reliably at 10 m. In addition, being dependent on external global datasets restricts the ability of government agencies to adjust class definitions, incorporate local expertise, or maintain long-term continuity across updates. For the MRNF, owning and controlling the production chain is therefore essential, not only to guarantee thematic adequacy and flexibility, but also to ensure reproducibility and stable annual updates.

To meet these operational needs, the Ministère des Ressources naturelles et des Forêts recently initiated the development of a dedicated 10-m land-cover product for Québec, the Cartographie de l'Occupation du Territoire Québécois (COTQ), produced from Sentinel-2 imagery using recent deep learning-based semantic segmentation. The methodological framework underlying the COTQ, its training data generation and architecture design, has been detailed elsewhere (Clabaut et al., 2024) and is therefore not the focus of this paper. Instead, the present work concentrates on an evaluation of the resulting product. This evaluation relies on two complementary perspectives: 1) a comparison with some major global 10-m LULC datasets, harmonized under a common legend allowing direct thematic and spatial assessment; and 2) a photo-interpretation campaign guided by disagreement maps.

Formal accuracy assessment remains a challenging aspect of land-cover validation: reference data must be independent, spatially representative, and interpretable with confidence, conditions that are difficult to satisfy simultaneously, particularly in heterogeneous environments (Moraes et al., 2024). Global products also differ in their class definitions, post-processing strategies, and ecological assumptions, making direct comparison nontrivial. Recent studies have reported substantial disagreement among global maps, including within boreal and temperate regions (Venter et al., 2022; Xu et al., 2024). For these reasons, evaluating a regional map such as the COTQ requires a combination of quantitative metrics, qualitative spatial analysis, and class-specific interpretation rather than relying on a single accuracy indicator. The goal of this study is therefore to assess how the COTQ performs relative to existing global products, and how faithfully it captures the ecological and anthropogenic patterns that are most relevant for annual monitoring of soil occupation and soil artificialisation across Québec.

Hence, the purpose of this study is not to introduce a new mapping methodology, but rather to evaluate the COTQ product as objectively, rigorously, and transparently as possible. Our assessment combines several complementary perspectives designed to determine how the COTQ compares to existing 10-m land-cover datasets and how reliable it is for operational monitoring. First, we evaluate the internal intrinsic consistency of the COTQ using metrics that capture spectral separability, object structure, and temporal stability. Structural indicators based on patch size, shape complexity, and object distribution have been widely used to characterize landscape patterns, although their interpretation remains sensitive to spatial scale, image resolution, and land-cover type (Masoudi et al., 2024). Second, we perform a direct comparison with the main global 10-m land-cover products (ESA WorldCover, Google Dynamic World and ESRI Land Cover), harmonized under a common legend, in order to quantify thematic agreement and identify in which ecological contexts each product performs better or diverges. This comparative analysis also allows us to identify the closest competing product for each land-cover type. Finally, we conduct a targeted expert photo-interpretation of sampled disagreement areas to determine which product is more frequently supported by very-high-resolution reference imagery. By combining these different angles of analysis, the study provides a comprehensive evaluation of the COTQ and helps clarify its position relative to modern global land-cover datasets, with a particular focus on its suitability for annual monitoring of soil artificialisation across Québec.

## 2. Background: Overview of the COTQ Production Pipeline

### 2.1 The COTQ taxonomy

The COTQ land-cover classes were selected to remain both ecologically meaningful and operationally robust at the 10-m spatial resolution of Sentinel-2, reflecting the MRNF's need for an annual, policy-oriented product capable of supporting province-wide monitoring of soil artificialisation. Several classes such as High Vegetation, Permanent Water Bodies, Bare Earth, and Exposed Rock, follow conventional physical land-cover definitions and align with international LULC taxonomies. However, other categories were adapted specifically for Québec's ecological structure and monitoring priorities.

Most global 10-m land-cover products include a single "Bare Soil" class that merges exposed rock and bare earth. In contrast, the COTQ explicitly separates these surfaces into two distinct classes, distinguishing bare earth from exposed rock. Because these substrates dominate different ecological domains (*i.e*. agricultural and disturbed soils versus Canadian Shield bedrock) and exhibit distinct spectral-structural behaviour, they were separated in the COTQ legend to avoid conflating fundamentally different environments.

Conversely, physiognomic types that cannot be reliably discriminated at 10 m, such as grasslands versus shrublands, were intentionally merged into a single Low Vegetation class. In boreal and subarctic landscapes, these vegetation types intermix at sub-pixel scales and exhibit substantial spectral overlap, making finer subdivisions unstable and of limited operational utility. A second important distinction from most global 10-m land-cover products is the absence of an explicit Crop class. This choice was intentional because it was judged more robust to represent crops strictly through their physical land cover (Low Vegetation or Bare Earth). This avoids imposing a land-use interpretation that cannot be reliably inferred from Sentinel-2 reflectance alone. Nevertheless, agricultural areas are of major interest for long-term land-use monitoring, and parallel efforts are underway to integrate a dedicated Crops class in future versions of the COTQ.

Three classes were absent from the taxonomy presented in Clabaut et al. (2024) and were introduced during the subsequent development of the provincial product: Moss & Lichen, to better represent the extensive cryptogamic cover of northern rocky environments; Wetlands, including peatlands, fens, shallow marshes, and wet meadows; and Ice/Snow, to explicitly identify transient frozen or snow-covered surfaces during individual Sentinel-2 inference dates. The Ice/Snow class, however, is not retained as a thematic land-cover class in the final annual product and is therefore not relevant to the comparative evaluation presented in this paper.

A unique feature of the COTQ legend is the three-tiered representation of urban environments, motivated by the need to monitor artificialisation with greater precision than is afforded by a single "built-up" class. Roads are mapped separately from other urban structures because major transportation corridors form long, coherent, and visually distinct linear features that remain separable from residential or industrial surfaces. Residential areas, dominated by detached houses, small streets, and mixed gardens/vegetation, form a characteristic spatial pattern at 10 m and warrant their own category to track low-density urban expansion. Industrial and commercial areas, including large warehouses, factories, commercial zones, and all forms of heavy land transformation such as mines, quarries, earthworks, construction sites, and large excavation areas, are grouped under an Industrial/Commercial class. Although these surfaces may be physically composed of bare soil or rock, they represent clear human-driven land modification. Incorporating them into the industrial urban class ensures that major anthropogenic transformations are correctly captured in artificialisation analyses.

This classification scheme therefore combines physical land-cover realism, functional land-use relevance (mining, earthwork…), and practical discriminability at 10-m resolution. It avoids unnecessary subdivisions where separability is unreliable, introduces categories critical for Québec's ecological and policy context, and provides an explicit structure for representing different forms of urbanization, ultimately supporting a clearer and more accurate assessment of land-cover change and soil artificialisation across the province.

The table below summarizes the different classes, their definitions, and the associated classification challenges.

*Table 1: COTQ taxonomy, identification and confusions challenges.*

| Class | Identification criteria | Associated confusions |
|---|---|---|
| **Residential areas** | Urbanized zones dominated by single-family housing, characterized by a heterogeneous texture mixing buildings, low vegetation, and tree cover. | Confusion with vegetation-bare soil mosaics due to high vegetation presence. Internal road networks are often difficult to distinguish. |
| **Commercial and/or industrial areas** | Areas composed of large buildings with very limited vegetation, often associated with extensive parking lots or storage yards. | Confusion with large residential complexes or other artificial surfaces where the distinction relates more to land use than land cover. Road networks may partially disappear. |
| **Roads** | Linear artificial surfaces with spectral properties close to bare soil but a distinctive elongated geometry. | Very difficult to detect in dense urban areas. Differentiation between highways, roads, paths, or minor lanes is generally not feasible at this resolution. |
| **High vegetation (Forest)** | Arboreal vegetation, tree cover. | Confusion with forest is rare. Most errors originate from shadows being misclassified as water, as well as from very smooth-textured surfaces being incorrectly labeled as low vegetation. |
| **Low vegetation** | Low and dense vegetation with a smooth and homogeneous appearance, visually comparable to grassland or shrubland. | Generally limited confusion. Occasional misclassification may occur with forest edges or with wetland areas dominated by low vegetation. |
| **Bare soil (earth type) with sparse vegetation** | Surfaces primarily composed of exposed soil but exhibiting sparse or discontinuous vegetation cover. This class is assigned when pixels initially classified as bare soil (earth type) exhibit an NDVI greater than 0.2, indicating the presence of vegetation. | Generally limited confusion. Frequently includes agricultural surfaces at different phenological stages. |
| **Bare soil (earth type)** | Non-vegetated surfaces, typically brown, with uniform or furrowed textures. | Includes agricultural fields prior to vegetation growth and recently cleared lands; confusion mainly related to land use. |
| **Bare soil (rock type)** | Exposed bedrock or rocky surfaces with a relatively smooth appearance, usually occurring in natural environments. | Confusion with excavation sites, construction areas, mines, or earthworks. Distinguishing between rock, crushed stone, and asphalt can be problematic. |
| **Bare soil (rock type) with sparse vegetation** | Rocky surfaces partially covered by sparse vegetation,. This class is assigned when pixels initially classified as bare rock exhibit an NDVI greater than 0.2, indicating non-negligible vegetation presence. | Confusion with moss and lichen dominated areas or mixed rock-vegetation mosaics, particularly in northern environments. |
| **Wetland** | Areas characterized by saturated soils, shallow water, or hydrophytic vegetation, including marshes, peatlands, and wet meadows, often with heterogeneous textures and transitional boundaries. | Confusion with low vegetation, forest edges, or shallow water bodies, particularly in riparian or seasonally flooded environments. |
| **Moss and lichens** | Cryptogamic vegetation typically occurring in northern or rocky environments, appearing as low, patchy vegetation | Separation becomes challenging when this class is intermixed with other vegetation types or sparsely distributed over rocky environments. |
| **Permanent water bodies** | Smooth surfaces with low reflectance, often dark, occasionally affected by specular reflections. | Confusion with shadows due to similar reflectance. Turbulent or aerated water may appear bright and be misclassified. |
| **Clouds** | Large objects with very high reflectance and diffuse boundaries. | Generally well detected. Semi-transparent clouds remain challenging as they alter surface radiometry while allowing partial visibility of underlying features. |

## 2.2 Data preparation and model training

Before presenting the comparative evaluation of land-cover products, it is essential to outline the method used to generate the MRNF's Carte d'Occupation du Territoire du Québec (COTQ). Although the primary aim of the present article is not to describe the full production workflow of the COTQ, understanding the principles behind its creation is crucial for interpreting its behaviour relative to WorldCover, ESRI, and Google products. The COTQ is derived from a semi-supervised annotation and training pipeline originally introduced in (Clabaut et al., 2024) for southern Québec and later extended to the entire province. This approach, based on automated region extraction, synthetic mosaic generation, and efficient sample selection, underpins the operational production of the MRNF land-cover map evaluated here. For clarity and completeness, we therefore provide below a concise update of the methodology from (Clabaut et al., 2024), as it forms the conceptual and technical foundation of the COTQ.

*2.2.1 Building the synthetic dataset*

The central contribution of (Clabaut et al., 2024) lies not in the choice of model architecture or training strategy, but in the development of an efficient semi-supervised annotation pipeline specifically designed for medium-resolution Sentinel-2 imagery. Manual pixel-level labeling at 10 m resolution is both imprecise and prohibitively time-consuming, and the study addressed this limitation by introducing a semi-automated sample-collection methodology that drastically reduces the need for human annotation.

The process begins with the application of the Felzenszwalb superpixel segmentation algorithm (Felzenszwalb and Huttenlocher, 2004) to RGB-NIR Sentinel-2 images. This produces an over-segmentation where each region is internally homogeneous with respect to spectral content and spatial structure, thereby providing annotation units. Instead of delineating complex boundaries manually, annotators simply select superpixels that unambiguously correspond to a given land-cover class. This transforms the annotation task from pixel-level labeling into a rapid selection procedure focused on a set of spatially consistent, low-noise training samples. Because each selected region captures an entire homogeneous surface, the approach dramatically reduces annotation time while improving label quality.

Once a sufficient collection of superpixels has been gathered across all targeted land-cover classes, these samples are recomposed into large synthetic mosaics (Figure 1). The mosaicking process preserves the original spectral signatures of the selected superpixels while arranging them into new, spatially diverse configurations. This synthetic-image generation step greatly expands the effective size and heterogeneity of the training dataset without requiring additional manual labeling. It also mitigates geographical bias by decoupling the spatial distribution of classes from the limited set of locations where annotations were originally collected.

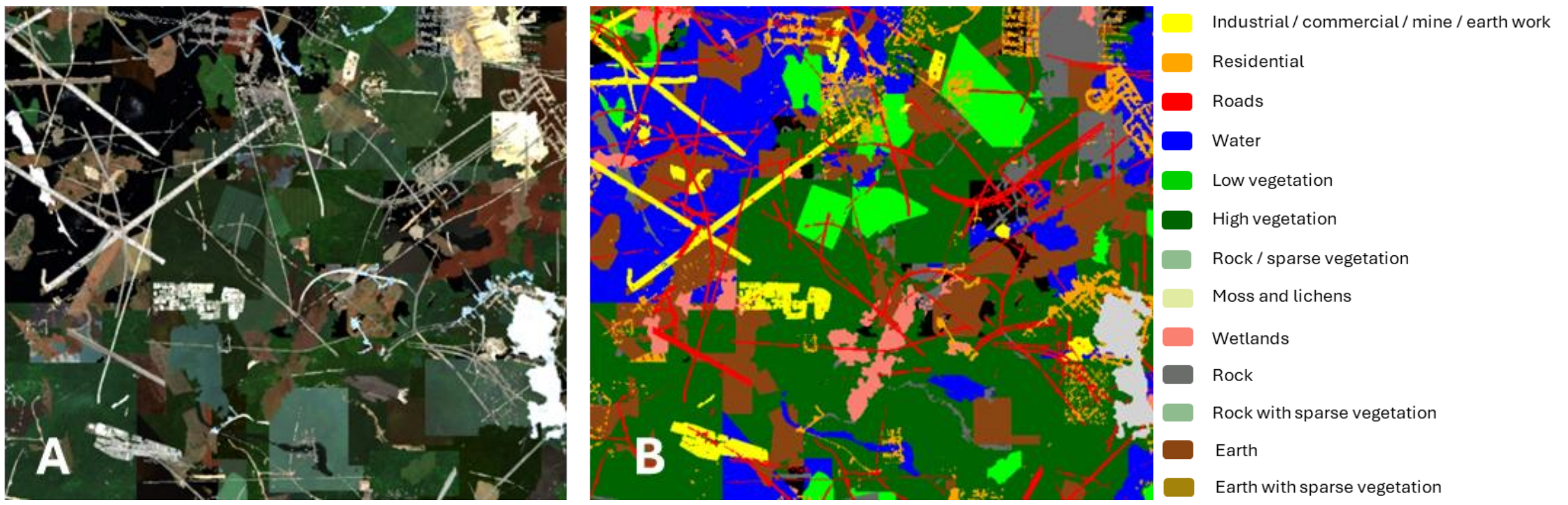

*Figure 1 : Example of synthetic training data. (A) Synthetic Sentinel-2 RGB mosaic generated by assembling numerous patches pre-segmented using Felzenszwalb's algorithm and selected to represent diverse land-cover conditions. (B) Corresponding synthetic ground-truth mask. These paired mosaics were*

*used to train the initial segmentation model without requiring manual pixel-level labeling of real Sentinel-2 imagery.*

The final model is trained on these synthetic mosaics, which combine real Sentinel-2 spectral information with artificially varied spatial contexts. The 2024 study systematically evaluated multiple semantic-segmentation architectures: DeepResUNet (Yi et al., 2019), DeepLabv3 (Chen et al., 2017), and SegFormer (Xie et al., 2021), to determine how effectively models could learn from such semi-supervised training data. Results showed that all three architectures benefited substantially from the synthetic sample-generation pipeline, but DeepResUNet consistently achieved the best performance, with notably higher accuracy and more stable predictions across spatial contexts. It is worth noting that, despite using Attention mechanisms (Vaswani et al., 2023), SegFormer lacked the spatial accuracy to delineate fine structures. Remarkably, all models, especially DeepResUNet, already demonstrated strong mapping capability before any fine-tuning on real images.

These findings confirm the central proposition of the method: that large-scale, high-quality Sentinel-2 training data can be produced from semi-automated pre-segmentation, the selection of semantically consistent and well-separated polygons, and mosaic generation, without requiring extensive human annotation. The performance gains arise from the abundance of clean and homogeneous class samples, the diversity of spatial configurations introduced through the mosaics, and the reduction of label noise that typically affects manual annotation at 10-m resolution. As a result, the study validated that the synthetic-guided pipeline is sufficient to train competitive deep-learning models for land-cover mapping while dramatically reducing manual effort. Before applying the temporal aggregation procedure described below, an important difficulty had to be addressed: early versions of the COTQ revealed that many classification artefacts originated from clouds and shadows rather than from the model itself. Thick clouds were generally well detected, but semi-transparent clouds produced large, systematic errors because the underlying land surfaces remained partially visible yet exhibited altered brightness and contrast from cloud shadows. In such conditions, forest canopies were frequently misclassified as Built-Up due to artificially brightened reflectance, whereas cast shadows often caused the model to assign Water to darkened forest cover. When a given acquisition contained extensive cloud cover, sometimes more than 50 % of the footprint, these misclassifications propagated into the subsequent majority vote (see section below). Although redundancy across dates helps suppress isolated errors, we observed that heavily clouded years could still accumulate enough wrong predictions (e.g., Water from shadows, Built-Up from cloud brightening) to bias the final composite.

To mitigate these issues, the model was adapted to ingest NDVI and NDWI alongside the original four Sentinel-2 reflectance bands (RGB-NIR). Although the scientific literature remains divided on whether deep neural networks truly benefit from explicitly providing such derived indices (Rajah et al., 2019; Phiri et al., 2020; Zhao et al., 2022; Lozano-Tello et al., 2023), since, in principle, convolutional architectures can learn relevant band relationships on their own, their inclusion here serves a different purpose. Rather than supplying additional “information”, the indices introduce illumination-invariant spectral cues that reduce the impact of brightness fluctuations caused by clouds, shadows, seasonal anisotropy, or varying atmospheric conditions. In practice, they help the model separate genuine surface differences from artefacts driven by reflectance magnitude, thereby improving temporal stability and class consistency without altering the underlying physical content of the input imagery.

#### 2.2.2 Data augmentation

Data augmentation is a key component of modern deep-learning workflows, as it improves generalisation and robustness by exposing models to diverse geometric, radiometric, and contextual perturbations of the training imagery (Lei et al., 2019). In remote sensing, augmentation is particularly critical for handling clouds, shadows, and seasonal variability, and recent studies have even injected synthetic atmospheric artefacts such as artificial clouds or haze to increase robustness to occlusions (Adedeji et al., 2022; Rad, 2024; Sierra et al., 2025). To the best of our knowledge, our pipeline is the first to use dedicated augmentation operators to simulate semi-transparent clouds (Figure 2) and cast shadows using real cloud rasters. These operators blend cloud reflectance onto training images and reassign the corresponding mask labels, forcing the network to learn the spectral signatures of clouds and shadows explicitly rather than confusing them with surface classes. After integrating these augmentations, the systematic artefacts nearly disappeared, and the per-date predictions became sufficiently stable for reliable temporal compositing. Each selected cloud patch was converted into a spatially varying transparency field (α, ranging from 0 to 1), derived directly from its reflectance structure. Pixels with low α behave as thin, translucent clouds through which surface texture remains partially visible, whereas high-α pixels emulate dense, opaque clouds. The augmentation blends cloud reflectance onto the base image using this transparency field while simultaneously updating the segmentation mask to reflect cloud presence. As a result, the network is explicitly exposed to realistic gradients of cloud opacity and the associated spectral distortions, preventing it from confusing bright cloud-veils with built-up surfaces or interpreting dark cloud shadows as water.

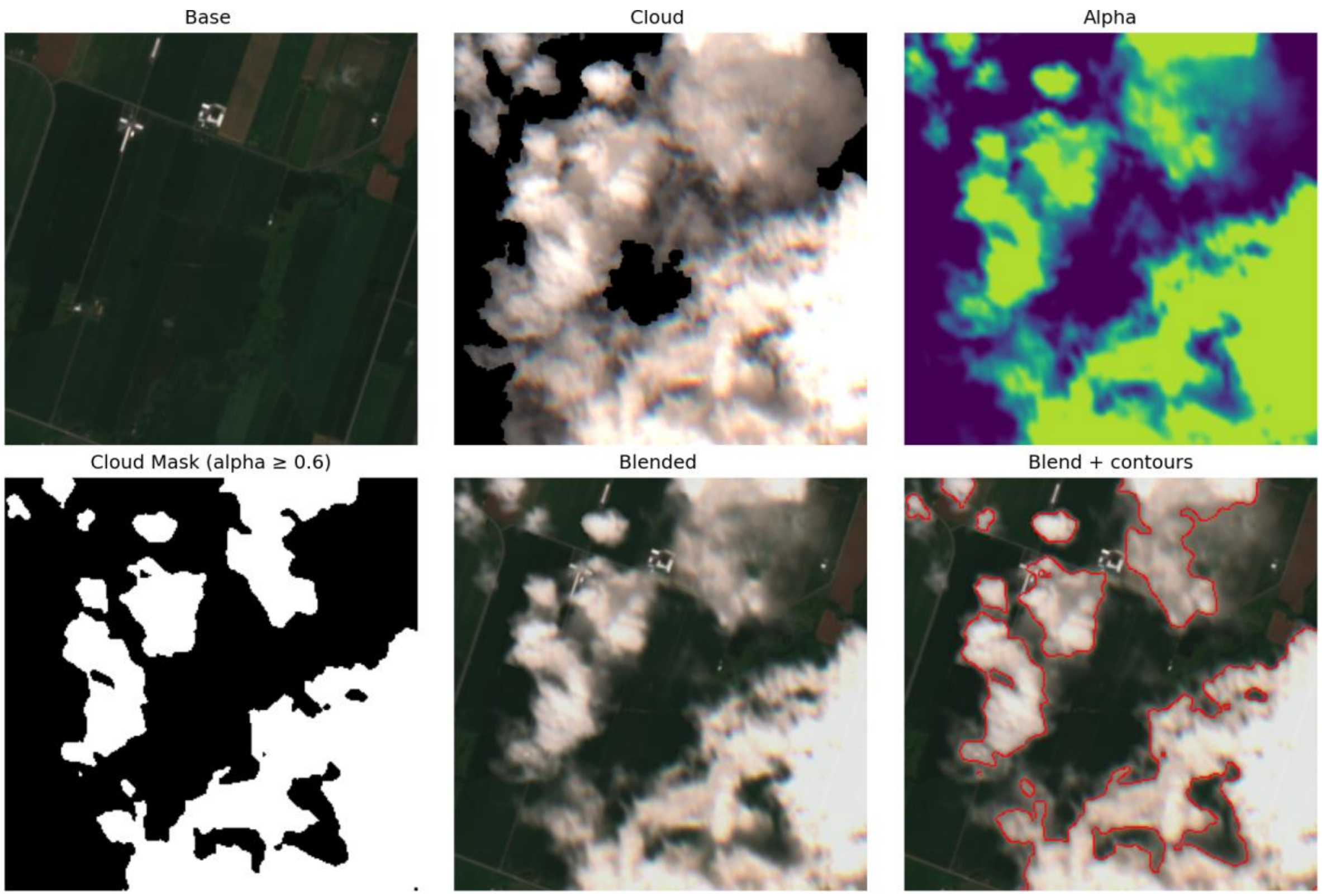

*Figure 2 : Simulation of semi-transparent cloud contamination used during data augmentation. A. Sentinel-2 base patch. B. Real cloud patch extracted from a high-resolution cloud dataset (RGB). C. Corresponding transparency field (α, 0-1) derived from cloud reflectance. D. Binary cloud mask obtained by thresholding α ≥ 0.6. E. Synthetic cloud-contaminated image obtained by alpha-blending the cloud patch over the base patch using the transparency field. F. Same composite image with the cloud-mask boundaries overlaid in*

*red; this visualization was used to verify that the augmentation produced realistic spatial patterns of semi-transparent clouds.*

This improved robustness is essential for the majority-vote framework described below: without reducing cloud- and shadow-induced misclassifications at the per-date level, temporal aggregation alone would not have been sufficient to ensure a consistent and ecologically plausible annual land-cover product.

### 2.3 Temporal aggregation and majority-vote compositing

Once the model had been trained, it was applied independently to each available Sentinel-2 image over the Québec Province. Only Sentinel-2 acquisitions acquired between May 15 and September 15 were considered for annual map production. This resulted in approximately 2800 cloud-filtered scenes (< 66 %), corresponding to about 8-11 usable acquisition dates per Sentinel-2 footprint and per year (Figure 3). Each inference produced a full 10-m land-cover map for a given date, expressed on the common legend used for the COTQ product.

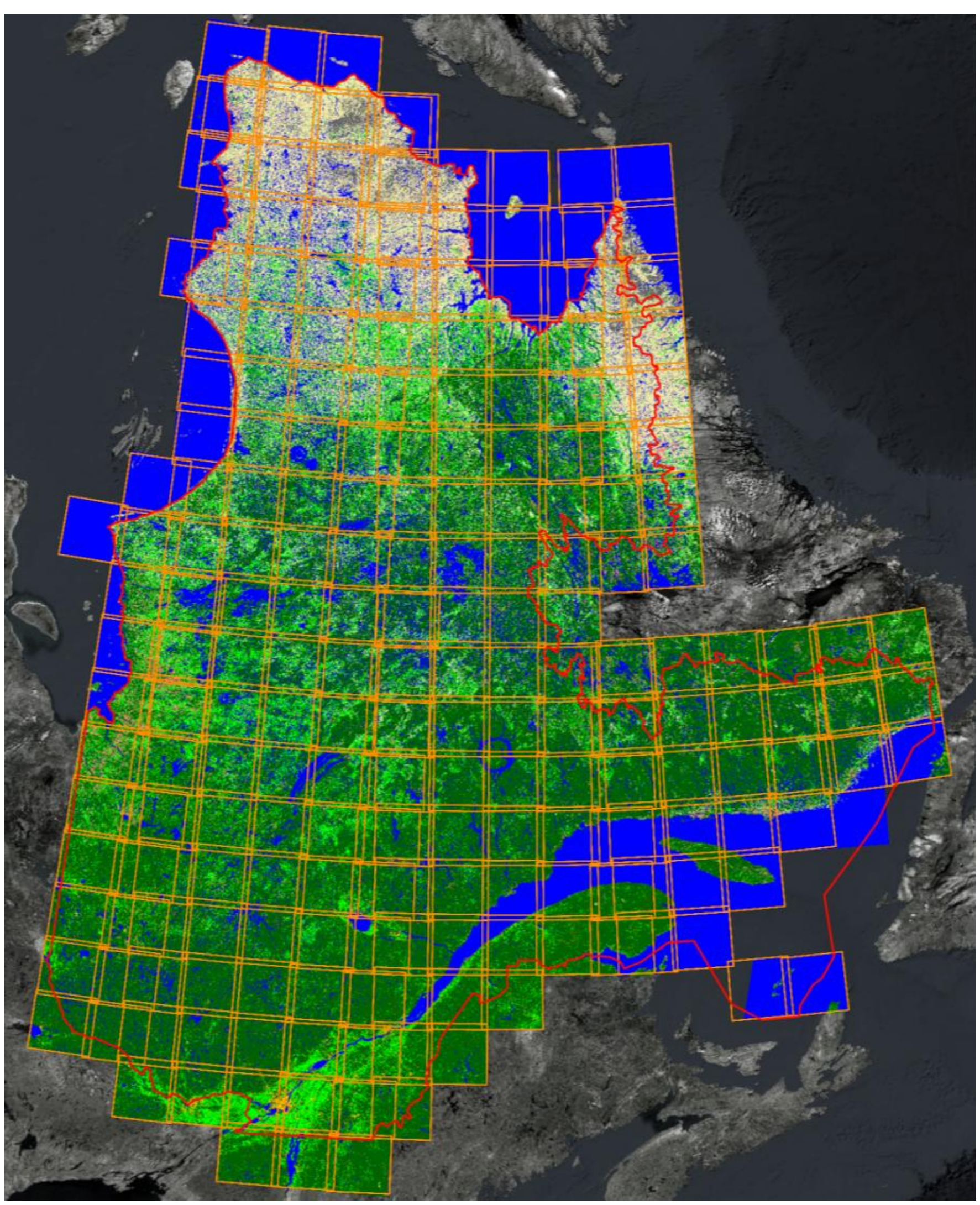

*Figure 3 : Spatial extent of the COTQ land-cover product over Québec. The red outline delineates the official boundary of the province, while the orange polygons represent the Sentinel-2 tile footprints used to generate the annual land-cover maps.*

Although all Sentinel-2 images were classified individually to preserve full temporal provenance, the per-date predictions were aggregated to produce a single temporally coherent annual map. Land-cover predictions can vary between acquisitions because of genuine intra-annual changes in surface conditions or land-cover state, seasonal phenological dynamics, and observation-related effects such as illumination differences, vegetation moisture, water-level variations, snow, clouds, and shadows. For each Sentinel-2 footprint and year, the available classifications were therefore stacked and combined using a pixel-wise majority vote. Predictions corresponding to NoData, Cloud, or Ice/Snow were excluded so that only valid surface observations contributed to the annual label.

The aggregation also incorporates limited information from the previous year to stabilize classes that are particularly prone to temporal ambiguity. Previous-year observations are retained only for Forest and Low Vegetation, thereby providing additional temporal context without allowing the complete previous-year map to constrain current land-cover changes. Additional rules are applied when current-year observations are nearly balanced (40–60%) for recurrent ambiguous situations. Wetland predictions can be reinforced by observations from the previous year, while ambiguities between Forest and Low Vegetation are resolved from their combined current- and previous-year frequencies. Repeated Bare Soil observations within such Forest-Low Vegetation ambiguities favour Low Vegetation, reflecting seasonal exposure of the ground rather than persistent forest cover. If the combined evidence remains exactly balanced, the initial majority-vote result is retained.

This temporal aggregation reduces isolated date-specific classification artefacts while preserving genuine land-cover changes and maintaining traceability to the original Sentinel-2 observations (Fig. 4).

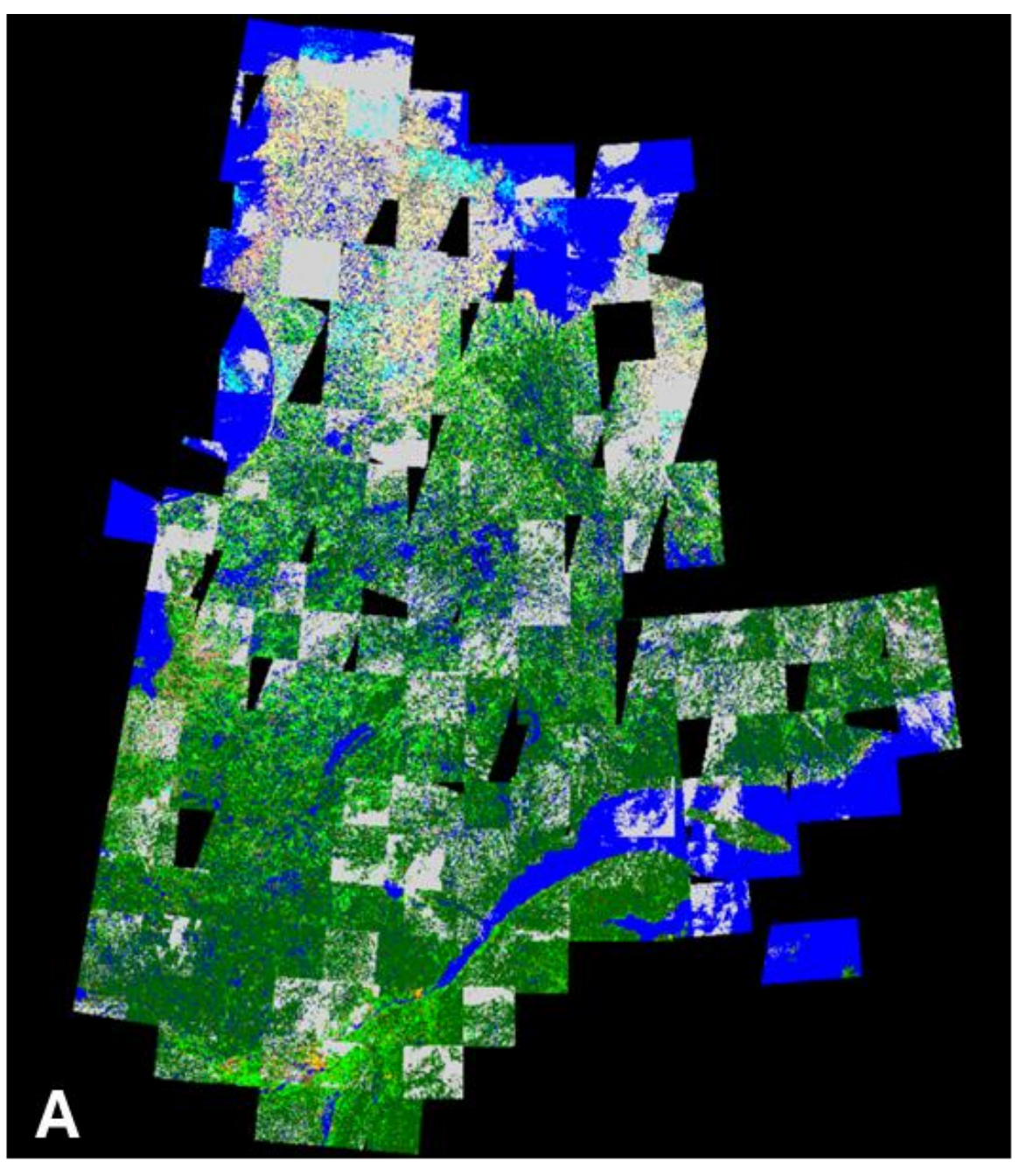

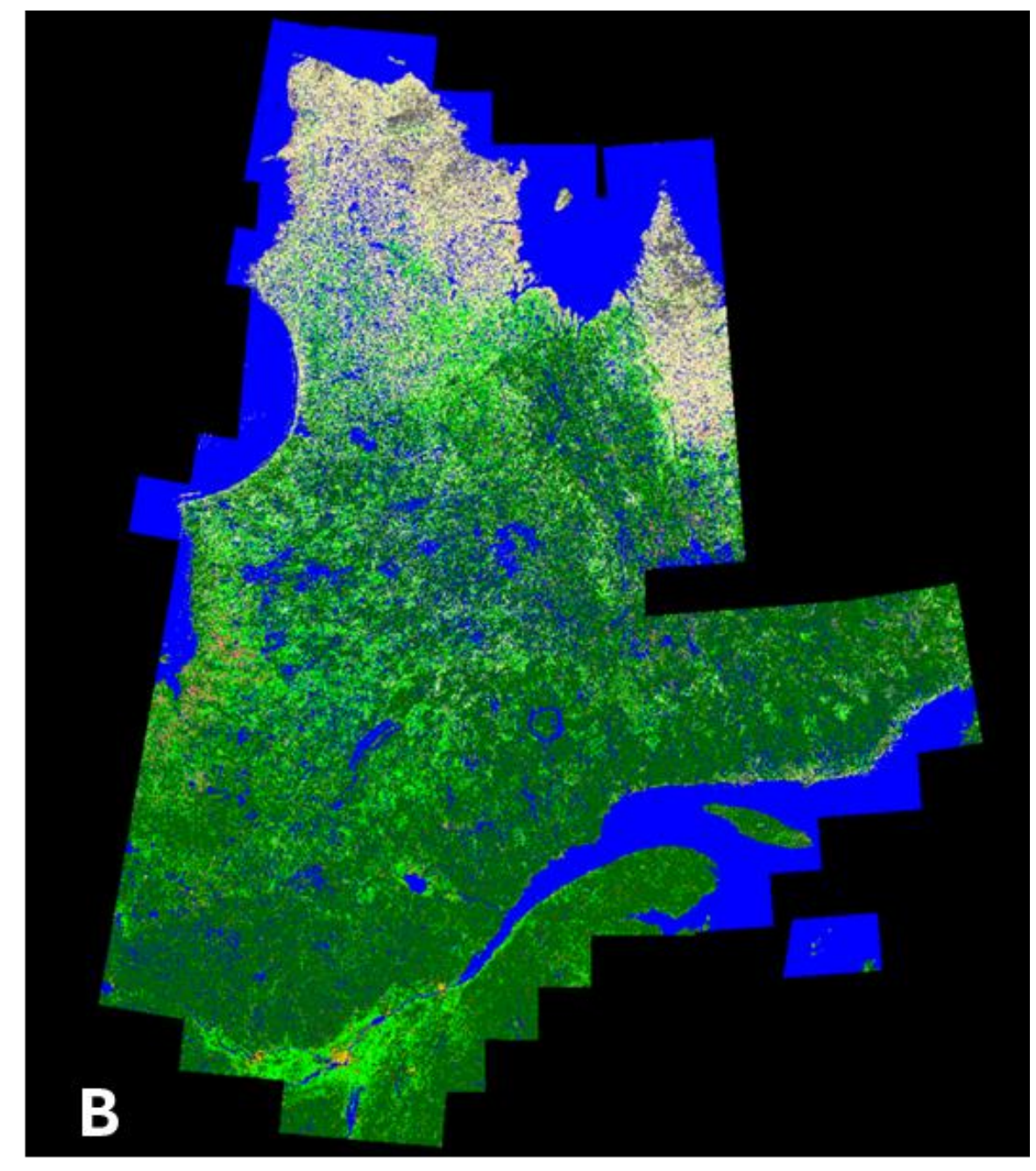

*Figure 4 : Example of the majority-voting process applied to Sentinel-2 predictions across the Québec Province. (A) Raw per-date semantic segmentation results, shown as a mosaic of all individual yearly predictions. Strong variability between dates is visible due to seasonal changes, clouds, and transient snow or water. No-data areas (in black) can occur because of partial orbits. (B) Final temporally-stabilized land-cover map after applying the majority-voting procedure. The resulting product is spatially coherent and suitable for provincial-scale land-cover mapping.*

### 2.4 Active learning process

Beyond the initial training phase, the COTQ production relied on an active learning procedure (Settles, n.d.) designed to iteratively correct systematic model errors. After each province-wide inference, the maps were visually inspected to identify recurring misclassification patterns. A particularly important issue concerned the confusion between urban areas and rocky or lichen-dominated surfaces, whose reflectance properties often overlap in Sentinel-2 imagery. Rather than attempting to correct all occurrences of this problem across Québec, a single representative site was selected for each error type. The principle was to correct only one well-chosen area and allow the next training cycle to generalize this correction to the rest of the territory. In practice, four such active-learning iterations were performed. For each cycle, the locally corrected masks were injected directly into the training dataset and re-used within the synthetic-mosaic pipeline, ensuring that the model repeatedly encountered, and learned to fix, its own typical failure cases. Although these improvements cannot be quantified against field truth, since no reference data exist at this spatial and geographic scale, they resulted in clear and systematic visual gains. One illustrative example is shown in Figure 5: the model initially failed to consistently separate anthropogenic surfaces from rocky and cryptogamic substrates in northern environments, but this error pattern was dramatically corrected after the very first active-learning iteration.

To make the corrections fast and precise at 10-m resolution, an interactive editing tool was developed in-house, as no existing software supported the required functionalities. Implemented in Python with GDAL and PyQt5, the tool displays a Sentinel-2 patch and its corresponding classification mask with a fully configurable color table. The operator first selects a patch (typically 512×512 pixels), then edits the mask using three complementary modes: (1) polygon replacement, which reassigns all pixels of selected source classes inside the polygon; (2) threshold-based editing, which uses NDVI, NDWI, band-reflectance or red/blue ratio thresholds within a polygon to modify only the pixels that satisfy both the spectral criterion and the specified classes; and (3) freehand line drawing with adjustable width, useful for correcting elongated or fragmented structures. All modifications are applied directly to the raster and can be undone through an internal history stack. Each corrected patch produces a pair of GeoTIFF files, image and updated mask, stored in a dedicated “manual correction” directory with preserved georeferencing. These patches are then incorporated seamlessly into the synthetic-mosaic generator, transforming targeted human feedback into additional supervision and enabling efficient active-learning refinement of the land-cover model.

Following the production of the first version of the southern Québec land-cover map, two additional classes, Ice/Snow and Moss & Lichen, were incorporated into the taxonomy to extend the product to the entire province and to better capture the ecological diversity of northern environments. In Version 1, cryptogamic surfaces such as moss and lichen were embedded within the broader Rock / Sparse Vegetation class due to their limited spatial extent in southern regions. However, these substrates become dominant across vast portions of the northern Shield, necessitating a dedicated class. Thanks to the semi-supervised sample-collection and synthetic mosaicking pipeline developed in (Clabaut et al., 2024), this expansion of the label set was achieved rapidly and with minimal manual workload. New representative samples for Ice/Snow and Moss & Lichen were simply collected from high-latitude Sentinel-2 footprints, inserted into the existing training database, and automatically re-mosaicked into synthetic training images alongside the original classes. This modular data-generation process allowed the taxonomy to evolve without re-annotating large datasets from scratch, enabling the efficient creation of subsequent versions of the provincial land-cover product.

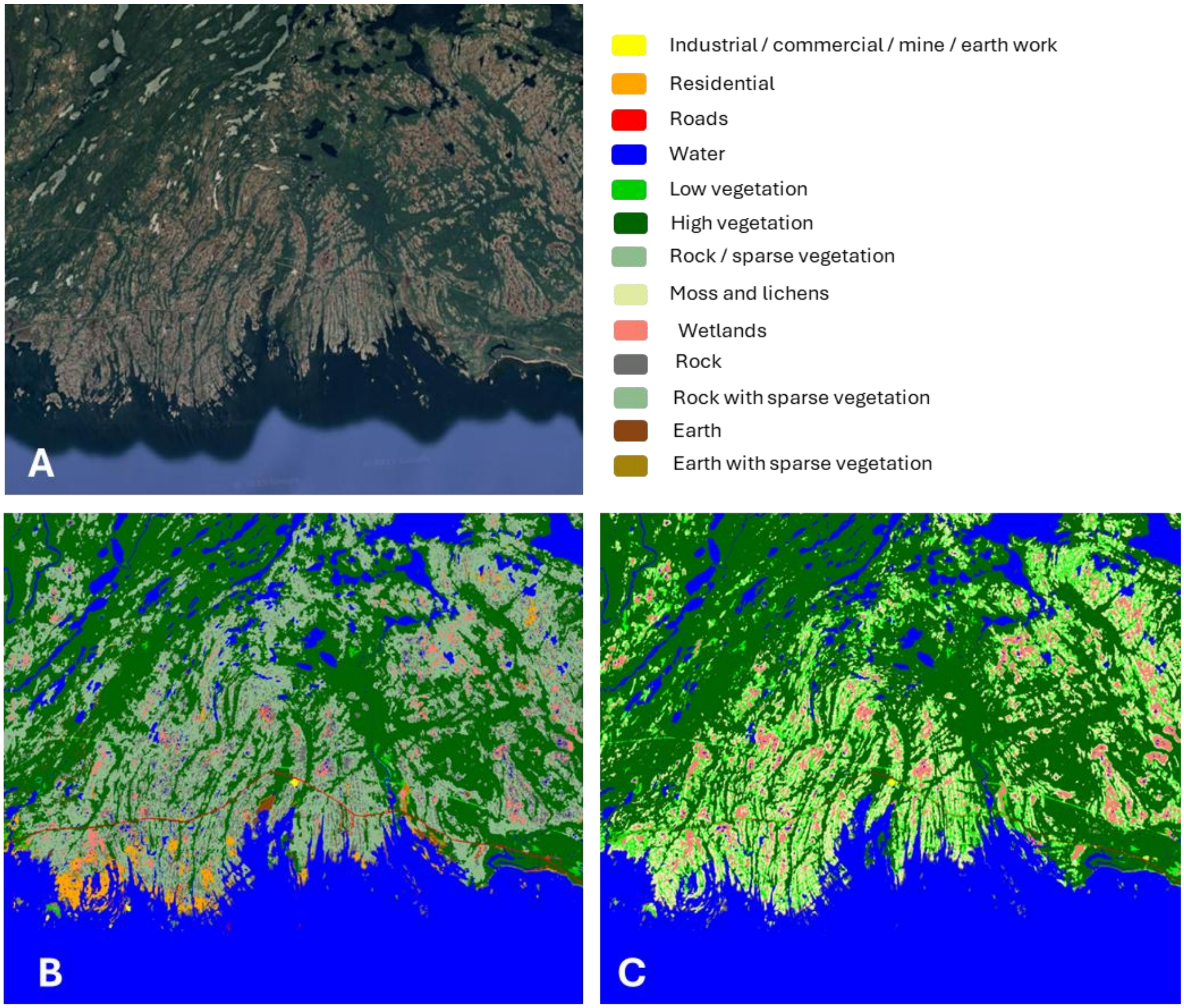

*Figure 5 : Improvements from active learning and the introduction of a dedicated moss-lichen class illustrated over the Wastishou Migratory Bird Sanctuary, a site located on the North Shore of Québec. (A) High-resolution reference imagery of the study area. (B) Classification from Version 1 of the MRNF land-cover model (before active learning and before introducing the moss-lichen class). In this version, rocky, cryptogamic, and sparsely vegetated substrates were grouped into a single "rock & sparse vegetation" class. Large areas of low vegetation and lichens are misclassified as "residential". (C) Classification after the active-learning workflow and after adding a dedicated moss-lichen class. Confusion between "residential" surfaces and low vegetation mixed with cryptogam-dominated substrates was strongly*

*reduced. The explicit separation of moss-lichen cover from bare rock also improves thematic detail and yields a more ecologically meaningful representation of this environment.*

The procedures described above define the complete operational workflow used to generate the annual COTQ product. The main stages of this production chain, including semi-supervised training-sample selection, synthetic mosaic generation, cloud–shadow augmentation, DeepResUNet training, iterative active learning, per-date Sentinel-2 inference, and temporal majority-vote compositing, are summarized in Figure 6.

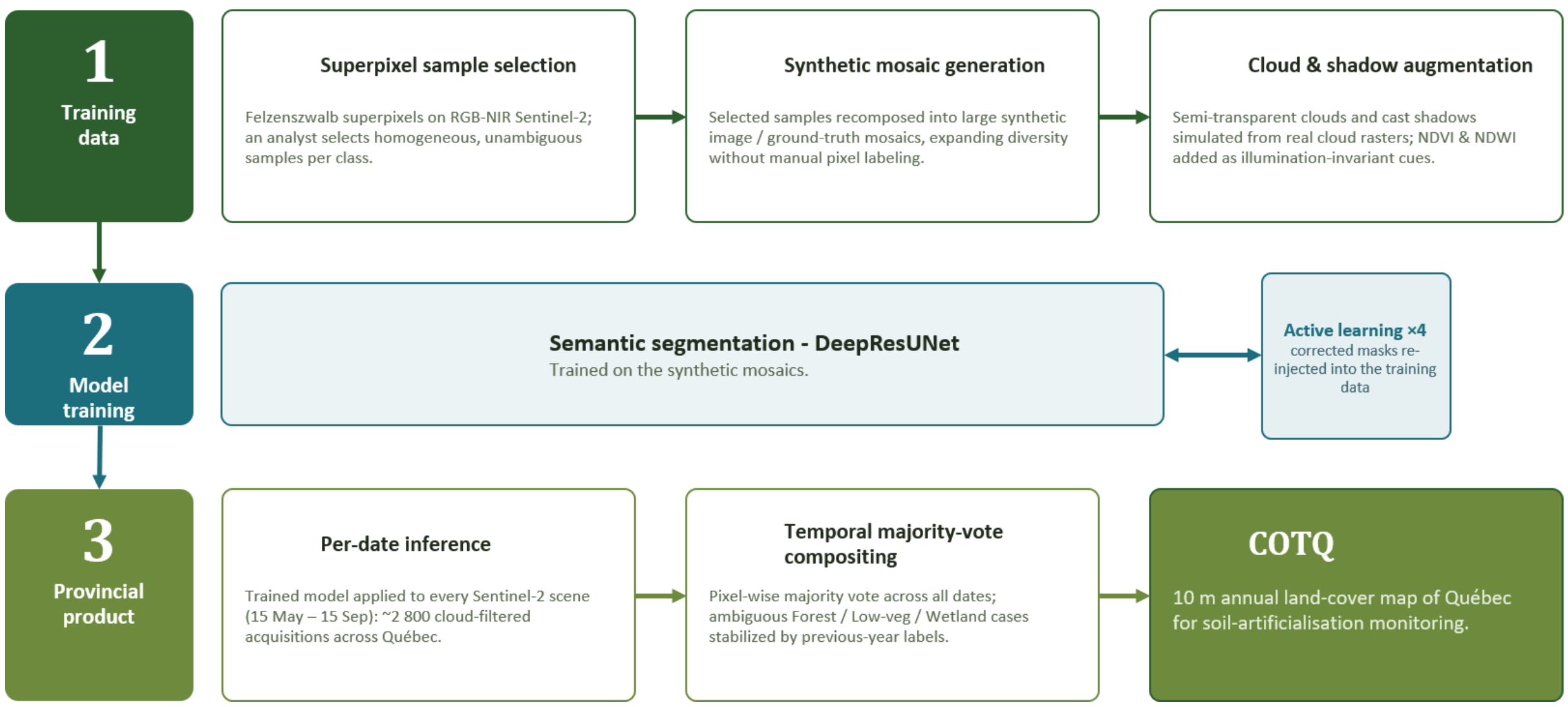

*Figure 6: Overview of the COTQ production workflow. The process is organized into three main stages: (1) training-data generation through superpixel-based sample selection, synthetic mosaic construction, and cloud–shadow augmentation; (2) DeepResUNet semantic-segmentation training, complemented by iterative active-learning corrections; and (3) province-wide production, including per-date Sentinel-2 inference followed by temporal majority-vote compositing to generate the final annual 10-m COTQ land-cover product for Québec.*

## 3. Materials and Methods for the COTQ Evaluations

### 3.1 Evaluation Framework in the Absence of Ground Truth

The objective of this study is to assess the behaviour, consistency, and operational suitability of the COTQ Land Cover classification product. As discussed previously, a major challenge is the absence of a reliable, large-scale ground truth. In this context, several well-established and independently evaluated land-cover products exist (ESA WorldCover, ESRI Land Cover, and Google Dynamic World). Although none can be considered an authoritative ground-truth representation of Québec, their published validation results make them meaningful external benchmarks for comparative evaluation. Accordingly, the evaluation strategy has two complementary goals: 1) Determine the relative consistency between products, in order to identify which existing product is closest to the COTQ classification; 2) Perform a targeted expert-based validation using photo-interpretation on high-resolution imagery, guided by discrepancy maps and focusing on the most meaningful areas of disagreement.

This section details the methodology employed to harmonize land cover legends, compute structural similarity metrics, perform inter-product comparison, measure within-class spectral separability, and finally conduct a photo-interpretation-based assessment of key disagreement regions.

3.2 Choice of Sentinel-2 Evaluation Areas

The eight Sentinel-2 footprints were deliberately selected to capture as much as possible the ecological and land-cover diversity of Québec (Figure 7), ensuring that the comparison between land-cover products is not biased toward a limited set of environmental conditions. Together, these footprints span the major bioclimatic domains of the province (*Classification écologique du territoire québécois*, 2021), from densely populated southern landscapes to the northern shield.

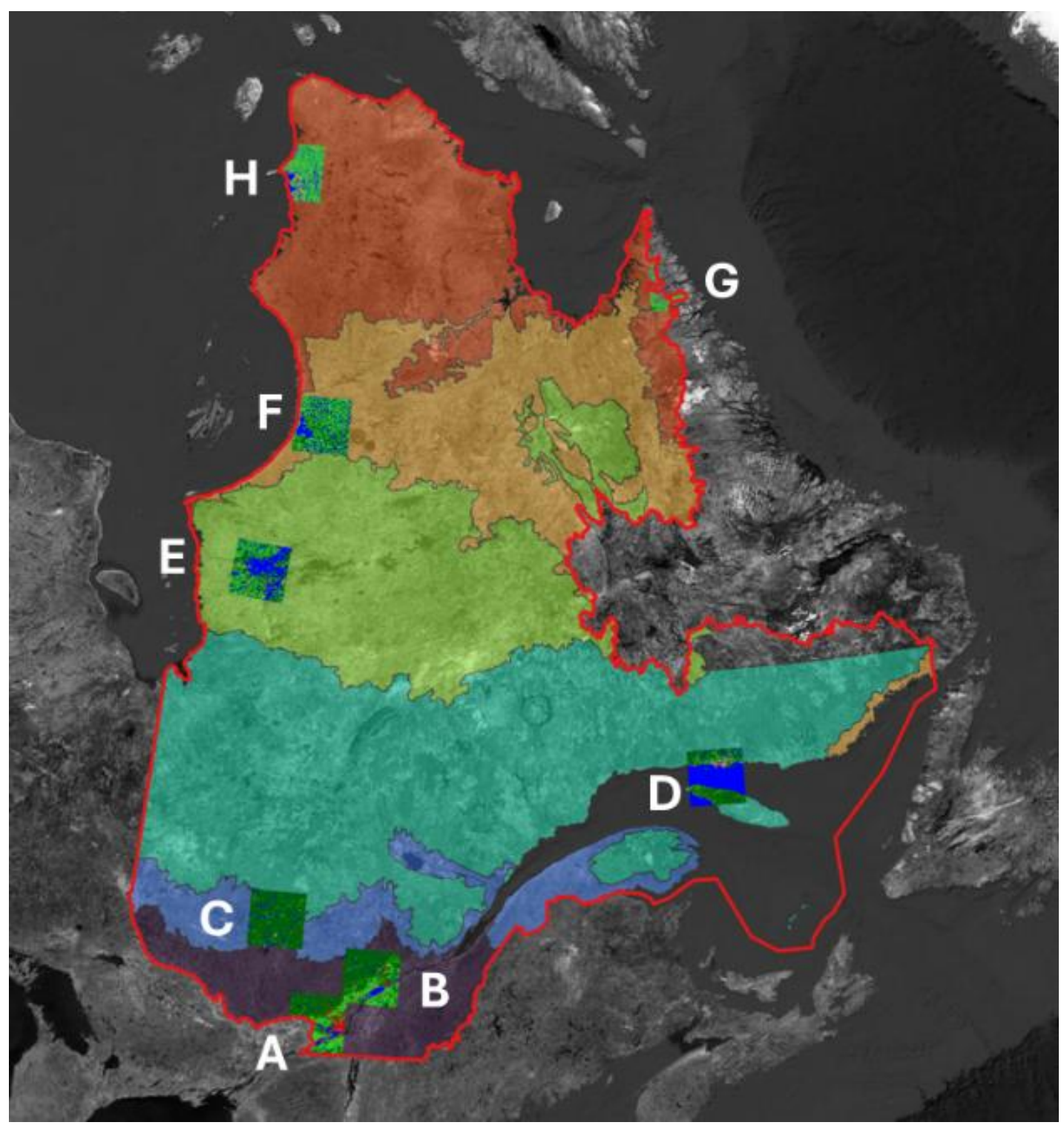

*Figure 7 : Location of the eight Sentinel-2 evaluation sites across the province of Québec. Footprints A and B (T18TWR and T18TXS) fall within the Deciduous Forest Domain. Footprint C (T18TVT) is located within the Mixed Forest Domain. Footprint D (T20UMA) lies within the Closed-Crown Boreal Forest Domain. Footprint E (T17UQV) is situated in the Open Boreal Forest. Footprint F (T18VVH) corresponds to the Forest-Tundra. Footprint G (T20VML) and H (T18UVN) are located within the Low Arctic Tundra ecozone.*

First, two evaluation sites A and B were chosen within the Deciduous Forest Domain, where urbanization, agricultural mosaics, and fragmented landscapes produce complex spatial patterns. Including these areas ensures that the evaluation captures conditions where built-up structures, fine-scale vegetation, and mixed land uses pose significant classification challenges. These southern footprints also provide the most realistic settings for assessing the representation of anthropogenic surfaces such as residential areas, industrial sites, or road networks. Evaluation area B is not redundant with footprint A despite both being located in the deciduous forest zone. B was specifically selected because it includes the extensive wetland complexes surrounding Lac Saint-Pierre, a Ramsar-designated site and one of the most ecologically significant wetland systems in Québec ("Ramsar Sites Information Service," 2001). Earlier versions of the COTQ land-cover mapping erroneously assigned the wetland area entirely to forest. Validating this footprint was therefore essential to ensure that the comparison between products adequately addresses the challenges associated with wetland detection, shoreline marshes, flooded meadows, and seasonally dynamic hydrological regimes, all of which are prominent in the Lac Saint-Pierre region.

A second evaluation site in the Mixed Forest Domain (C) offers an intermediate setting where coniferous and deciduous vegetation co-occur alongside human-modified areas. This region provides a transition zone, testing the ability of products to handle gradual ecological gradients.

The evaluation sites D and E were selected within the Closed-Crown and Open Boreal Forest Domains, which together cover a substantial portion of Québec's landmass. These regions contain large, relatively homogeneous forest stands interspersed with wetlands, lakes, river corridors, and disturbances such as logging or wildfires. The inclusion of these domains allows the assessment of classifier behaviour in environments where the dominant challenge lies not in urban complexity but in distinguishing among vegetated surfaces, hydrological features, and natural disturbances.

Finally, three evaluation areas located in the Taiga and Low Arctic Tundra (F, G, H) were included to represent Québec's northern landscapes. These environments are characterized by extensive lichen-moss carpets, exposed bedrock, sparse shrub vegetation, wetlands, and periglacial features. Because these environments differ from those in the temperate south, their inclusion is essential for evaluating each product's robustness across highly contrasting ecological conditions. They are also crucial for examining known challenges such as the detectability of mosses and lichens.

The combination of southern, mid-latitude, boreal, and arctic environments ensures that the selected evaluation areas jointly capture the major ecological gradients of Québec while also representing the land-cover situations most relevant to COTQ applications, including urban expansion, transportation corridors, wetlands, industrial and mining landscapes, and northern rocky environments. The objective was therefore to maximize ecological diversity, thematic class representation, and landscape heterogeneity within a limited number of large Sentinel-2 footprints. This diversity strengthens the generalizability of the conclusions and enables a more nuanced understanding of where and why different products diverge. Because the ESA WorldCover product has not been updated since its 2021 release, this year was chosen as the common temporal reference for all comparisons, ensuring temporal consistency across datasets.

### 3.3 Semantic harmonization of Land Cover Classes

The four land cover products involved do not share the same nomenclature. Some datasets present fine-grained subclasses, whereas others merge several vegetation types. Before any comparison, all products were remapped to a harmonized legend composed of the following classes inherited from the COTQ classification: 3 - Forest, 4 - Low vegetation, 6 - Rock/Bare, 7 - Built-up, 9 - Water, 10 - Wetland and 11 - Ice/Snow. The numbers correspond to the class on the actual raster map. The purpose of this harmonization is not to evaluate the original thematic legends themselves, but to enable comparison within a common operational framework aligned with the objectives of the COTQ. Consequently, some thematic distinctions present in the original products are intentionally simplified, and the results should be interpreted as a comparison of end-use suitability under a shared legend rather than a direct comparison of the original classification systems.

The complete mapping rules for each product are presented in Table 2.

*Table 2: Harmonization of land-cover classes across datasets to the COTQ legend.*

| | **Original classes** | **Final Classes** |
|---|---|---|
| **COTQ** | 3, high vegetation | 3, high vegetation |
| **ESA WorldCover** | 10, tree cover | |
| **ESRI Land Cover** | 2, trees | |
| **Google Dynamic World** | 1, trees | |
| **COTQ** | 4, low vegetation | 4, low vegetation |
| **ESA WorldCover** | 20, shrubland<br>30, grassland<br>40, crops<br>100, moss and lichens | |
| **ESRI Land Cover** | 5, crops<br>11, rangeland | |
| **Google Dynamic World** | 2, grass<br>4, crops<br>5, shrub and scrub | |
| **COTQ** | 6, bare soil / rock | 6, bare soil / rock |
| **ESA WorldCover** | 60, bare / sparse vegetation | |
| **ESRI Land Cover** | 8, bare ground | |
| **Google Dynamic World** | 7, bare | |
| **COTQ** | 1, industrial / commercial<br>7, main roads<br>8, residential | 7, built-up, urban areas |
| **ESA WorldCover** | 50, built-up | |
| **ESRI Land Cover** | 7, built area | |
| **Google Dynamic World** | 6, built area | |
| **COTQ** | 9, permanent water bodies | 9, permanent water bodies |
| **ESA WorldCover** | 80, permanent water bodies | |
| **ESRI Land Cover** | 1, water | |
| **Google Dynamic World** | 0, water | |
| **COTQ** | 10, wetlands | 10, wetlands |
| **ESA WorldCover** | 90, herbaceous wetland | |
| **ESRI Land Cover** | 4, flooded vegetation | |
| **Google Dynamic World** | 3, flooded vegetation | |

### 3.4 Inter-Product Structural Comparison

The first part of the analysis evaluates the spatial and structural similarity between classifications, independently of spectral information. The goal is not to measure "accuracy" but to characterize:

- how similar the products are in terms of internal consistency,
- which one is closest to the COTQ segmentation,
- and how each product represents object size, spatial detail, and boundary complexity.

Three metrics were used: object-size distributions (OSD), shape complexity, and Adjusted Rand Index (ARI).

*3.4.1 Object-Size Distribution (OSD)*

For each product and for each harmonized class, connected-components labeling is used to extract individual spatial objects. Their sizes (in pixels) are aggregated into histograms.

A product that generates a large number of small, fragmented objects indicates a finer-grained segmentation pattern, which may reflect either meaningful added detail or increased classification noise. A product producing large, oversmoothed objects indicates under-segmentation. Statistical comparison of OSD curves across products highlights systematic structural differences.

OSD is particularly useful because it characterizes the "style" of a classification independently of the spectral content.

*3.4.2 Shape Complexity (Compactness Measure)*

Because detailed, highly structured classes (urban, wetland, rocky coastlines) are expected to present irregular shapes, higher complexity scores indicate better discrimination of fine spatial structures. This complements the OSD: a product may produce objects of correct size but with overly regular shapes.

To quantify the intricacy of objects, we compute a compactness / complexity index for each spatial object:

$\mathrm{Comp}(S) = P(S)2/(4\pi A(S))$Where $A(S)$ is the area of the object and $P(S)$ is its perimeter.

Lower values indicate simpler object shapes, whereas higher values correspond to objects with more complex boundaries.

*3.4.3 Adjusted Rand Index*

The Adjusted Rand Index (ARI) is a widely used measure of agreement between two image partitions (Rand, n.d.). Unlike pixel-based accuracy metrics that rely on an external ground truth, the ARI evaluates the structural similarity between two classifications by examining whether pairs of pixels are assigned to the same or different classes in both maps. Importantly, the ARI includes a correction for chance agreement, making it particularly robust when classes are imbalanced or when several categories occupy only small portions of the scene. It is defined as follows:

Given two image partitions U and V, ARI is defined as:

$$\mathrm{ARI} = \sum_{ij} \binom{n_{ij}}{2} - \frac{\sum_i \binom{a_i}{2} \sum_j \binom{b_j}{2}}{\binom{n}{2}} \bigg/ \frac{1}{2}\left[\sum_i \binom{a_i}{2} + \sum_j \binom{b_j}{2}\right] - \frac{\sum_i \binom{a_i}{2} \sum_j \binom{b_j}{2}}{\binom{n}{2}}$$

where $n_{ij}$ is the number of pixels assigned to class i by the first map and class j by the second, $a_i = \sum_j n_{ij}$ and $b_j = \sum_i n_{ij}$ are class marginals for each classification and $\binom{n}{2}$is the total number of pixel pairs.

The Adjusted Rand Index (ARI) ranges from $-1$ to 1 and quantifies the level of structural agreement between two classifications. An ARI value of 1 indicates perfect agreement, meaning that the two segmentations are identical in terms of object partitioning. An ARI value close to 0 reflects an agreement no better than would be expected by random assignment, while negative ARI values indicate a level of correspondence worse than random alignment.

### *3.4.4 Intersection over Union*

The Intersection-over-Union (IoU), also known as the Jaccard Index (Jaccard, 1901), is a widely used metric in remote sensing to quantify the agreement between two categorical maps. For a given class c, IoU measures the overlap between the areas where both products assign class c, relative to the total area assigned to that class by either product. Formally, for two classifications A and B, the IoU for class c is defined as:

$$\mathrm{IoU_c} = \frac{|\, \mathrm{A_c} \cap \mathrm{B_c} \,|}{|\, \mathrm{A_c} \cup \mathrm{B_c} \,|} = \frac{\text{True Positives}}{\text{True Positives} + \text{False Positives} + \text{False Negatives}}.$$

An IoU score of 1 indicates perfect agreement (complete spatial overlap), whereas a score of 0 indicates no overlap at all.

Unlike pixel accuracy, which can be misleading when classes are imbalanced, IoU penalizes both overestimation and underestimation of class extents, making it particularly suitable for land cover comparisons. A high IoU between COTQ and another product means that both maps predict the same class labels in a spatially coherent manner.

### 3.5 Spectral Separability Metrics

Because structural similarity does not guarantee spectral consistency, a second family of metrics evaluates how each classification organizes Sentinel-2 reflectance data in multispectral feature space.

We assess spectral separability using three complementary indicators: 1) the Silhouette score, 2) the Calinski–Harabasz (CH) index, and 3) the Spectral Angle Mapper (SAM).

These metrics are derived exclusively from Sentinel-2 reflectance values rather than from spatial structure, providing a complementary view of the spectral organization induced by each classification. Although the analysis relies on only four Sentinel-2 bands (B, G, R, NIR), whereas the deep-learning models may also exploit spatial context, texture, and neighbourhood patterns, these metrics provide objective diagnostic indicators of class compactness and separability in a common spectral feature space. They should therefore not be interpreted as accuracy metrics, but as complementary measures of the spectral coherence associated with each classification.

For each of the eight evaluation areas, spectral samples were extracted from the summer 2021 Sentinel-2 RGB-NIR image selected for minimal cloud contamination. The same Sentinel-2 image and spatial extent were used to evaluate all four land-cover products within a given area. For the Silhouette and Calinski–Harabasz analyses, pixels assigned to NoData, Cloud, or Ice/Snow were excluded for each product, and up to 20,000 remaining valid pixels were randomly sampled per evaluation area and product. When fewer than 20,000 valid pixels were available, all were retained. No class balancing was applied, so the relative abundance of the harmonized classes within each sample reflected their occurrence in the corresponding mapped area. The 20,000-pixel limit was introduced primarily to keep the computational cost of the Silhouette calculation manageable.

Before computing the Silhouette and Calinski–Harabasz metrics, a single standardization transform was fitted to all valid Sentinel-2 pixels within each evaluation area and then applied identically to the spectral data used for all four products. Each band was therefore centred to zero mean and scaled to unit variance within the corresponding evaluation area. For SAM, spectra were instead L2-normalized because the metric is based on angular rather than Euclidean distance. Up to 20,000 valid spatial locations were randomly sampled once within each evaluation area, and the same initial locations were used for all four products before applying product-specific class filtering.

All spectral metrics were first calculated independently for each evaluation area. Regional results were subsequently aggregated using an unweighted arithmetic mean across the eight areas, so that each ecological region contributed equally to the reported summary values rather than in proportion to its number of valid pixels. Class frequencies were not artificially equalized during either sampling or aggregation. The resulting spectral metrics are therefore interpreted as descriptive measures of how the different land-cover products partition the same regional Sentinel-2 spectral spaces, rather than as estimators of province-wide spectral properties.

*3.5.1 Silhouette Coefficient*

The Silhouette Coefficient is a widely used cluster-validity metric that quantifies how well individual samples are grouped within their assigned class compared to other classes (Rousseeuw, 1987). For a given pixel $x_i$ belonging to class $C$, two quantities are computed:

- $a(i)$: the mean intra-class distance. The average distance between $x_i$ and all other points belonging to the same class $C$. It measures how compact the class is around that pixel.
- $b(i)$: the mean nearest-class distance. The smallest average distance between $x_i$ and samples from any other class.
  It measures how separated the pixel is from the closest alternative class.

The silhouette score of $x_i$ is:

$$s(i) = \frac{b(i) - a(i)}{\max(a(i), b(i))}$$

The global score is the average over all sampled pixels.

A high Silhouette value, close to 1, indicates that a pixel is well embedded within its assigned class, reflecting strong intra-class compactness and clear separation from neighboring classes. A Silhouette value around zero suggests that the pixel lies near a class boundary, where spectral signatures overlap and class separation is weak. Negative Silhouette values indicate that the pixel is, on average, closer to another class than to its assigned one, which points to a potential misclassification or to an inadequately defined class structure.

In the context of multispectral land-cover mapping, the Silhouette Coefficient evaluates the spectral validity of each class by jointly assessing intra-class compactness and inter-class separation at the pixel level. Because it measures, for every sample, how strongly it belongs to its assigned class relative to the nearest alternative class, the Silhouette score provides a sample-wise assessment of spectral coherence across the classification. Unlike the CH index, which summarizes global between- and within-class variance, the Silhouette formulation captures local class consistency, boundary ambiguity, and the extent to which individual pixels lie near or across class transitions.

*3.5.2 Calinski-Harabasz Index*

The Calinski-Harabasz (CH) Index (Calinski and Harabasz, 1974), also known as the Variance Ratio Criterion, evaluates the quality of a partition by comparing the dispersion between classes to the dispersion within classes. For a classification of N samples into K classes, the CH index is defined as:

$$\mathrm{CH} = \frac{\frac{\mathrm{B}}{\mathrm{K}-1}}{\frac{\mathrm{W}}{\mathrm{N}-\mathrm{K}}} \; \frac{\mathrm{B}}{\mathrm{K}-1}\Big/\frac{\mathrm{W}}{\mathrm{N}-\mathrm{K}}$$

Where B represents the between-class dispersion, defined as the sum of squared distances between each class centroid and the global centroid, weighted by class size, and W represents the within-class dispersion, defined as the sum of squared distances between each sample and the centroid of the class to which it belongs.

Interpretation:

- High CH value → classes are well separated and internally compact; between-class variance is large relative to within-class variance.
- Moderate CH value → classes show some structure, but separability is limited; spectral clusters may partially overlap.
- Low CH value → classes are poorly separated or highly dispersed; substantial spectral overlap or weak class definition.

Thus, larger CH values indicate sharper class boundaries and greater spectral compactness, while lower values correspond to overlapping or poorly defined classes. In the context of multispectral land-cover mapping, the CH index quantifies how well the reflectance distributions of the classes are separated in feature space. Since it relies exclusively on spectral distances rather than spatial patterns, it complements the Silhouette score by emphasizing global separability rather than the sample-wise consistency captured by the silhouette formulation. Combined, the two indices provide a broader view of the spectral validity of the classification independently from ground-truth labels, which is essential in studies where no authoritative reference map is available.

*3.5.3 Spectral Angle Mapper (SAM)*

The Spectral Angle Mapper (Kruse et al., n.d.) quantifies how similar two reflectance spectra are by measuring the angle between their spectral vectors:

$$\theta(\mathrm{x},\mathrm{y}) = \arccos\left(\frac{\mathrm{x}^{\top}\mathrm{y}}{\parallel \mathrm{x} \parallel \parallel \mathrm{y} \parallel}\right)$$

Interpretation

- $\theta = 0$: identical spectral signature
- Large $\theta$: different spectral shapes

In the context of multispectral land-cover mapping, the Spectral Angle Mapper (SAM) quantifies class discriminability by evaluating the angular similarity between spectral vectors. Because SAM is insensitive to illumination intensity, atmospheric shading, and overall brightness variations, it isolates the shape of the reflectance spectrum rather than its magnitude. This property makes SAM particularly well suited for environments where illumination varies strongly across the landscape, for example in rugged terrain, northern environments with shallow solar angles, or surfaces composed of bright minerals and pale substrates.

## 3.6 Photo-Interpretation

### *3.6.1 Discrepancy map*

A pixel-wise discrepancy map was generated to highlight disagreement between the COTQ Land Cover classification and the selected comparison product. This map identifies all pixels for which the two classifications differ (COTQ ≠ Selected Product). However, direct use of this map is not ideal for expert evaluation because a substantial fraction of the disagreement originates from thin class boundaries, where small geometric misalignments or resampling artefacts produce visually irrelevant 1 to 2 pixels strips of difference. These border effects do not reflect genuine thematic disagreement and would bias the qualitative assessment.

To mitigate this issue, a 3-pixel binary erosion was applied to the disagreement mask. This morphological operation removes all narrow disagreement ridges while preserving larger, spatially coherent disagreement patches. The resulting mask contains only meaningful, interpretable areas, which greatly improves the reliability of a human photo-interpretation. However, because the urban class is highly fragmented and composed of very fine spatial structures, applying morphological erosion would undesirably remove many meaningful disagreement pixels. For this reason, a dedicated qualitative and quantitative evaluation of the urban class was carried out prior to the erosion step. To quantify agreement with an independent high-resolution reference for the Built-Up class, a 50-cm reference classification derived from 2022 Jilin-1 imagery was used. A careful visual inspection confirmed that no significant urban development or demolition occurred between 2021 and 2022 over the study area, making this dataset a reliable high-resolution proxy ground truth. All land-cover products were resampled to 50 cm and compared to this reference using the Intersection over Union (IoU), focusing exclusively on the urban class. This combined morphological filtering, qualitative interpretation, and high-resolution quantitative comparison enables a robust evaluation of urban classification performance while avoiding bias due to trivial geometric disagreements.

### *3.6.2 Random Sampling of Disagreements*

For each Sentinel-2 evaluation site, only the disagreement areas remaining after the three-pixel erosion are used for evaluation. From these eroded disagreement regions, forty points per Sentinel-2 footprint are randomly sampled. Each sampled point is then manually inspected using Google Earth VHR imagery. The photo-interpretation was performed by a single analyst with more than 10 years of experience in remote sensing, including professional private-sector experience in aerial-image interpretation, semantic segmentation, and LiDAR point-cloud classification. During interpretation, neither the COTQ nor the WorldCover classification was displayed. Each sampled location was first assigned a reference land-cover class from the very-high-resolution imagery alone. Only after this thematic decision was made were the product labels in the attribute table consulted to record whether the interpretation supported COTQ, WorldCover, neither product, or an indeterminate outcome. The latter occurs in several situations: when the interpreter's expertise is insufficient to confidently assign a class; when the available verification imagery lacks the spatial detail necessary to resolve the land-cover object; or when the surface is partially obscured by ice, haze, or cloud cover. In these cases, the true label cannot be reliably established, and such points were recorded as indeterminate and excluded from the class-based quantitative comparison. The recorded interpretation outcomes were subsequently used to summarize the relative support for COTQ and WorldCover within the sampled disagreement areas.

The overall methodology is summarized in figure 8.

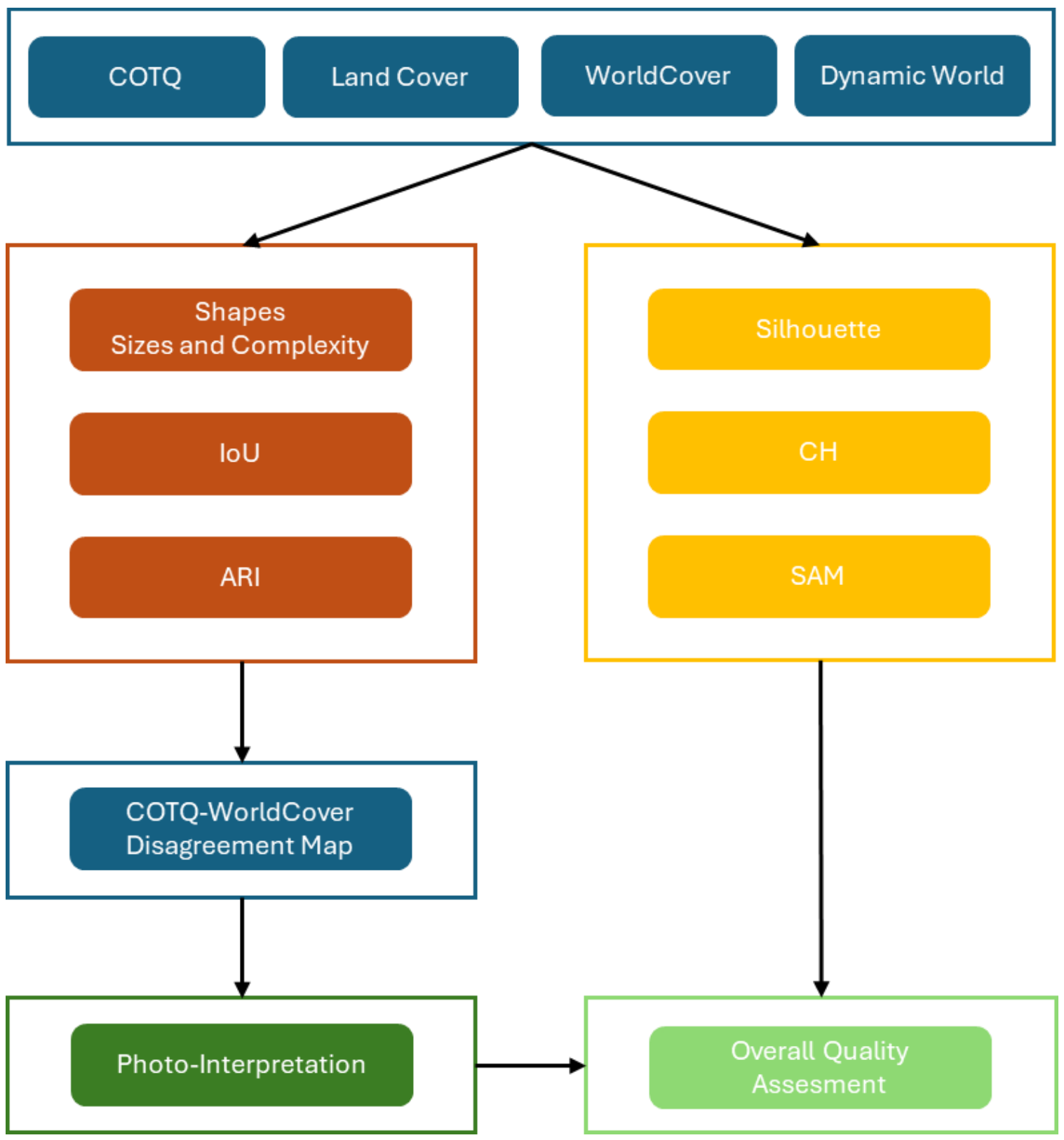

*Figure 8 : Overview of the evaluation framework. Four land-cover products (COTQ, ESA WorldCover, ESRI Land Cover, and Google Dynamic World) are assessed using two families of metrics: (1) object-based indicators (shape size distribution, shape complexity, IoU, ARI) and (2) spectral separability metrics (Silhouette, Calinski-Harabasz, SAM). A focused COTQ-WorldCover disagreement analysis is then performed, followed by targeted photo-interpretation and an overall quality assessment.*

## 4 Results and discussion

In this section, we present the qualitative and quantitative results for the different processus presented in the section 3. We also discuss these results and put them in perspective of the main goal of this paper, an objective evaluation of the COTQ classification.

### 4.1 Inter-Product Structural Comparison

#### *4.1.1 Object-Size Distribution (OSD)*

Computing the histograms of classified object sizes and boundary complexities provides a quantitative basis for interpreting the visual differences between the products. Figure 9 illustrates the distributions for water and urban areas. Two main patterns emerge. Google Dynamic World and ESRI Land Cover show very similar behaviour, with markedly fewer small objects compared to ESA WorldCover and the COTQ product. Conversely, WorldCover and COTQ contain fewer mid-sized objects than Google and ESRI. This indicates that WorldCover and COTQ products share a comparable object-size distribution, a trend confirmed in Figure 9, where smaller objects appear more frequently in both products. Visual inspection supports this observation: small lakes, narrow roads, and isolated buildings that are visible in the landscape are often absent from the ESRI and Google classifications.

Beyond these descriptive differences, the log-log regressions provide an additional comparative characterization of object-size distributions. WorldCover and COTQ exhibit stronger linear trends in log-log space ($0.79 \leq R^2 \leq 0.98$) than ESRI Land Cover and Dynamic World. Such approximately linear behaviour is descriptively consistent with a Zipf-like or heavy-tailed organization of object sizes, although the regressions are not intended as formal statistical tests of a power-law distribution. The stronger Zipf-like signature observed in WorldCover and COTQ indicates that their object-size distributions are more consistent with the hierarchical size structure expected for natural landscape elements. In contrast, the weaker linearity observed for ESRI Land Cover and Dynamic World is consistent with the stronger spatial generalization identified independently from the object-size histograms, boundary-complexity analysis, and visual examination of the products. The log-log analysis is therefore interpreted here as a comparative structural indicator rather than as evidence that any product follows a specific theoretical probability distribution.

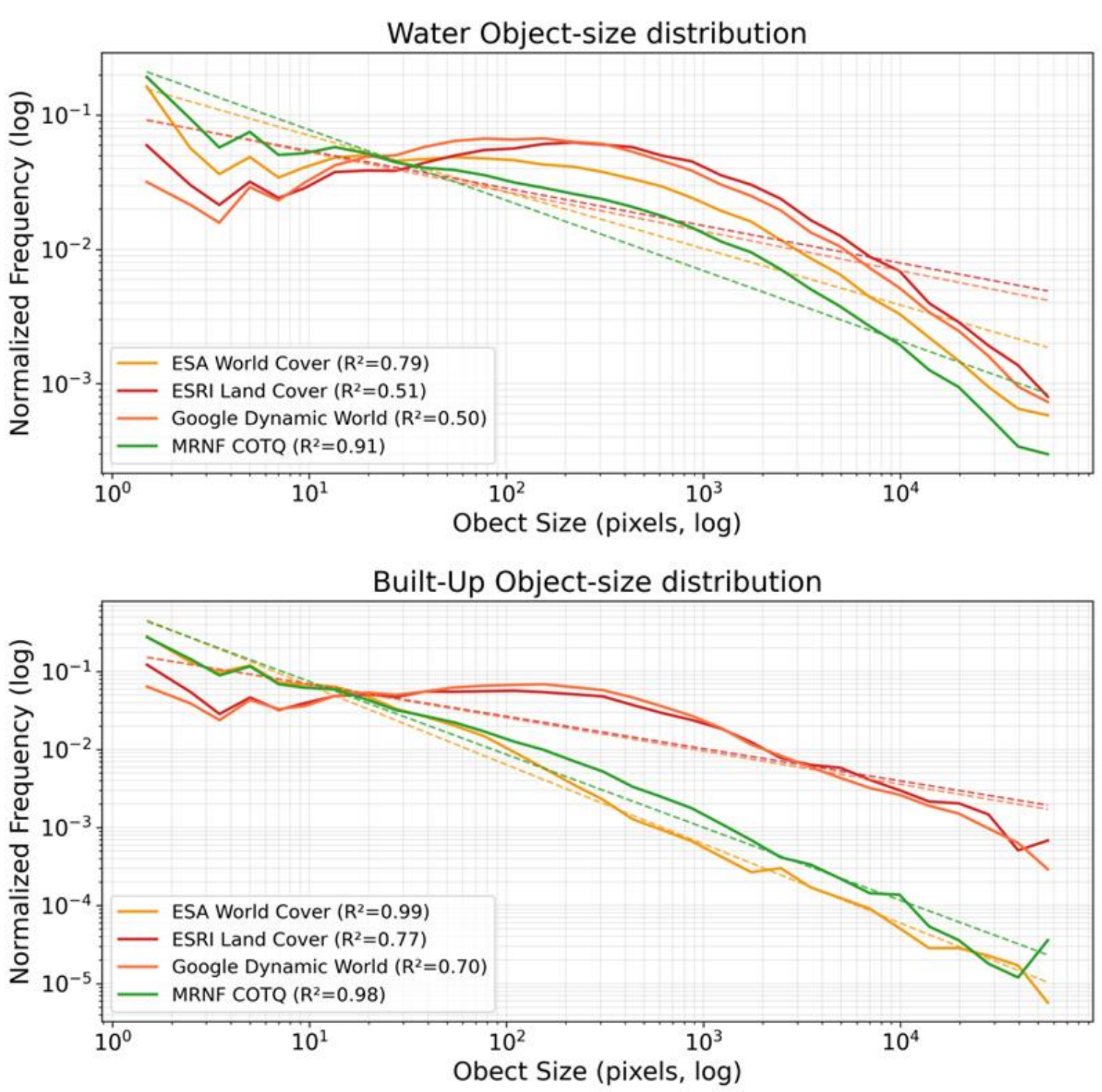

*Figure 9 : Log-log distributions of object sizes for the Water class (top) and Built-Up class (bottom) across ESA WorldCover, ESRI Land Cover, Google Dynamic World, and COTQ. Solid lines represent the empirical normalized frequency of object sizes, while dashed lines represent linear regressions fitted in log-log space. The coefficient of determination ($R^2$) reported in the legend describes the degree of linearity of each empirical distribution in log-log space and is used here as a comparative structural indicator. Higher $R^2$ values are interpreted as a stronger Zipf-like signature, not as formal evidence that the underlying object-size distribution follows a power law.*

*4.1.2 Shape Complexity*

The results on shape complexity are fully consistent with the segmentation pattern revealed by the object-size histograms: ESRI Land Cover and Dynamic World exhibit similarly simplified, coarse-grained delineations, whereas WorldCover and COTQ both produce finer, more fragmented geometries. Figure 10 presents the boundary-complexity values for each class and each product. Again, COTQ and WorldCover exhibit very similar behaviour, whereas Dynamic World and ESRI Land Cover tend toward simpler geometries. For the built-up class, crucial for distinguishing natural from artificial surfaces, the complexity values for COTQ and WorldCover are approximately five times higher. This difference aligns with the observation that ESRI and Google appear to apply a post-classification generalization (possibly erosion-dilation or similar smoothing), which removes small objects and isolated pixels while reducing boundary irregularity. Such a process would simultaneously explain the scarcity of small structures and the reduced shape complexity in these two products.

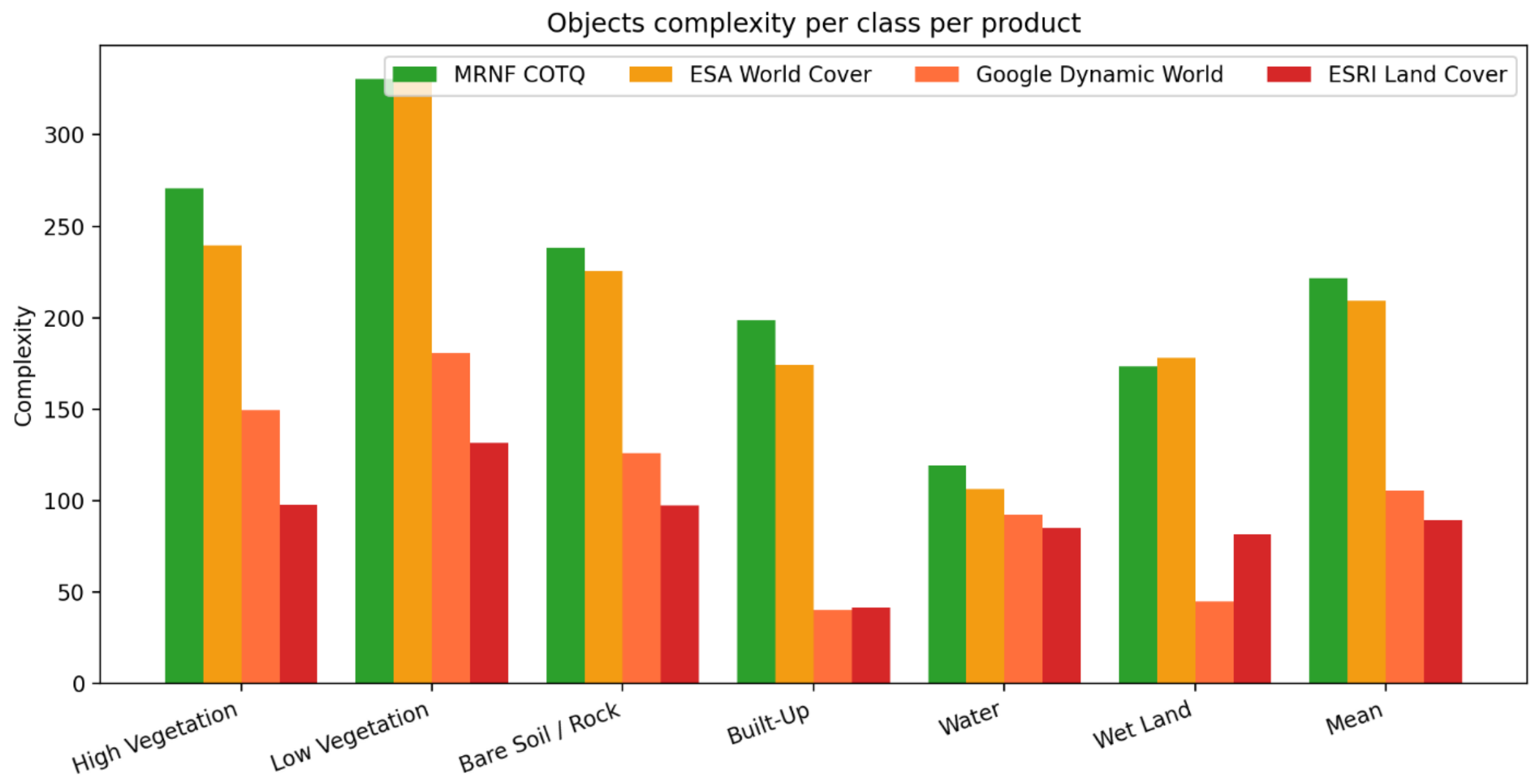

*Figure 10 : Comparison of object-boundary complexity across products. WorldCover and COTQ produce more intricate object geometries, whereas ESRI Land Cover and Google Dynamic World tend to generate smoother, more generalized shapes.*

*4.1.3 Built-Up class: localized reference-based assessment*

These indicators suggest that WorldCover and COTQ could preserve finer spatial detail (Figure 11). However, greater detail may reflect either improved depiction of heterogeneous environments or the presence of classification noise. In several locations, visual assessment confirms that at least part of this fine-scale detail corresponds to real cartographic features.

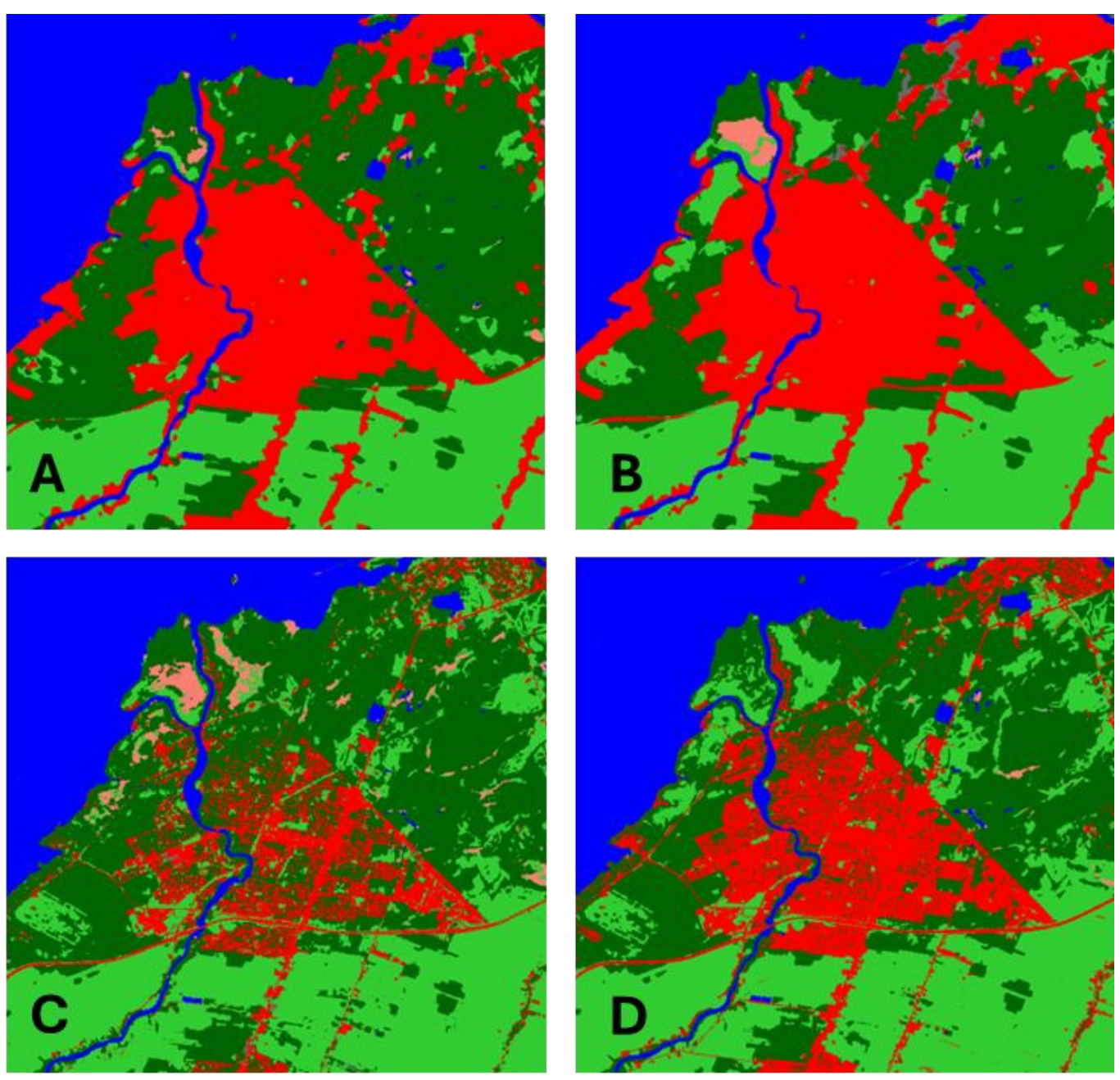

*Figure 11 : Comparison of land-cover classifications over an urban and peri-urban area. Panel A: Dynamic World. Panel B: ESRI Land Cover. Panel C: WorldCover. Panel D: COTQ. Differences in the delineation of built-up areas and vegetation illustrate the contrasting segmentation behaviors of the four products.*

To support this interpretation, we examined a sub-area (Figure 12) of the T18TWR Sentinel-2 footprint containing dense industrial and commercial zones as well as residential neighborhoods and compared the COTQ and WorldCover products with a 50-cm Jilin classification from 2022. This 50-cm reference classification was generated from 2022 Jilin-1 optical imagery using a DeepResUNet model trained with the synthetic-data generation strategy described in Clabaut et al. (2024). It was used here as a localized very-high-resolution reference for the Built-Up comparison, rather than as a province-wide ground-truth dataset. No significant land-cover changes occurred between 2021 and 2022, and although the 50-cm product is not formally assessed here, it is treated as a highly reliable reference relative to the 10-m Sentinel-2 resolution. Using this reference, IoUs were computed for the urban class. Google Dynamic World and ESRI Land Cover are not shown in the figure 12, as the analyses presented earlier demonstrated that their urban delineations differ fundamentally from those of WorldCover and COTQ, both in object size distribution and in shape complexity, making a direct visual comparison less informative in this specific context.

Although WorldCover and COTQ products show similar complexity and object-size patterns, visual inspection using high-resolution imagery suggests that COTQ more closely matches the structure of built-up areas, whereas WorldCover tends to underestimate urban surfaces, particularly in neighbourhoods where vegetation is abundant in streets or backyards.

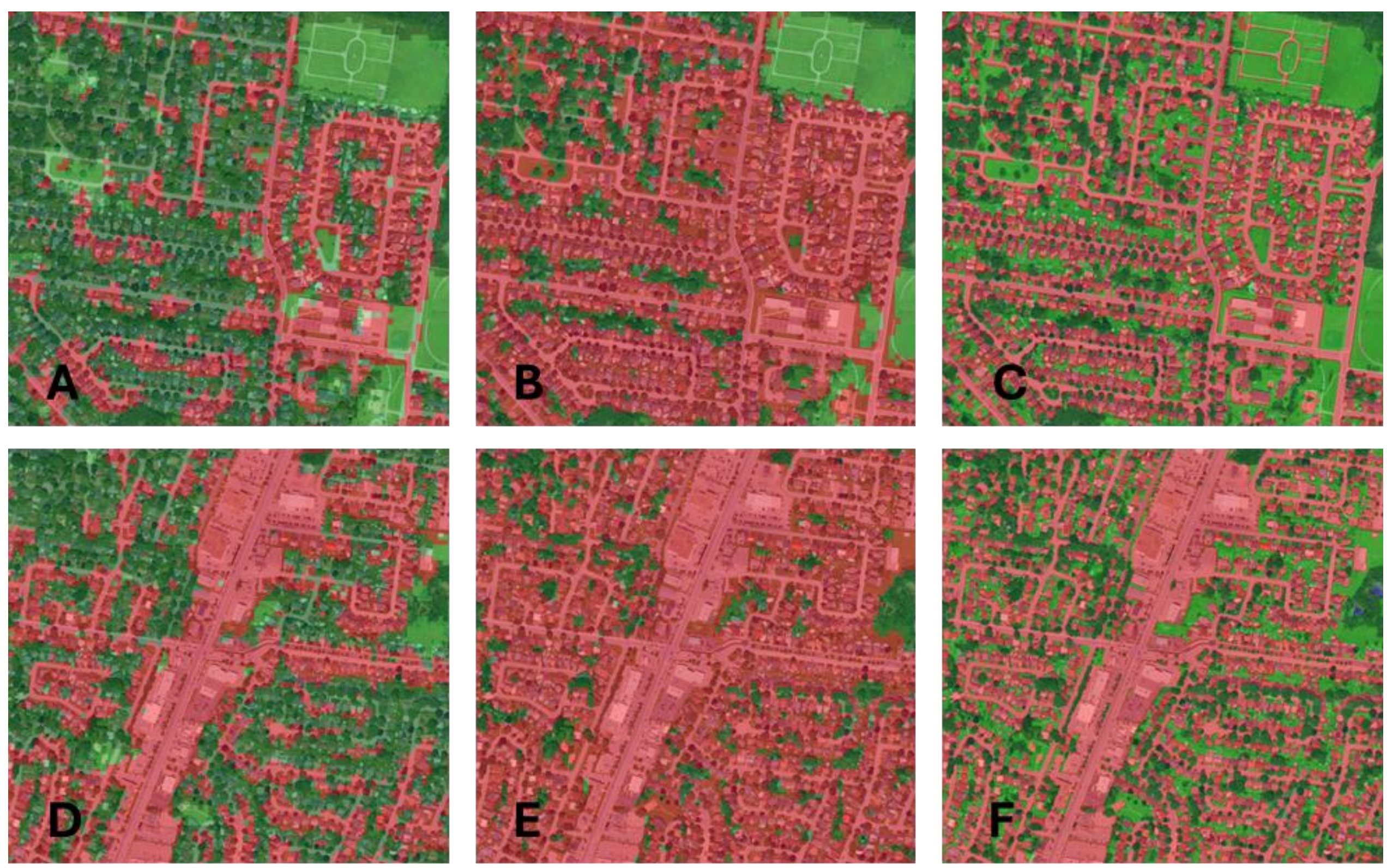

*Figure 12 : Comparison of built-up delineation in suburban neighbourhoods using a 50-cm reference classification (C and F) and two 10-m land-cover products. Panels A and D show ESA WorldCover; panels B and E show COTQ; panels C and F display the 50-cm Jilin classification considered as reference. Differences highlight the contrasting abilities of the two products to capture fine-scale built-up structures and high-low vegetation and urban mixtures.*

The IoU results corroborate these observations: COTQ achieves 67.5% agreement with the 50-cm reference, compared to 52.5% for WorldCover (Table 3). Higher spatial detail does not always guarantee higher accuracy, as shown by the fact that Dynamic World (57.4%) and ESRI Land Cover (55.2%) outperform WorldCover despite being more generalized. The combination of preserved fine structures and better alignment with the reference imagery explains why COTQ attains the highest IoU among the compared products.

*Table 3: IoU values relative to the 50 cm reference classification.*

| | **IoUs %** |
|---|---|
| **COTQ** | 67.5 |
| **WorldCover** | 52.5 |
| **Dynamic World** | 57.4 |
| **ESRI Land Cover** | 55.2 |

*4.1.4 Adjusted Random Index and Intersection over Union*

In addition to the analyses based on object-size distributions and shape-complexity indicators, two complementary agreement metrics, Adjusted Rand Index (ARI) and pairwise mean Intersection over Union (mIoU), were computed to quantitatively assess how closely each land-cover product aligns with the COTQ classification. The objective is to determine which external product provides the most coherent representation relative to COTQ, and is therefore the most appropriate candidate for the subsequent photo-interpretation assessment.

The ARI results (Figure 13) reveal clear structural similarities among the products. The strongest agreement is observed between COTQ and WorldCover, with an ARI score of 0.75. This value stands out relative to the other pairwise comparisons, which remain in the 0.67-0.68 range.

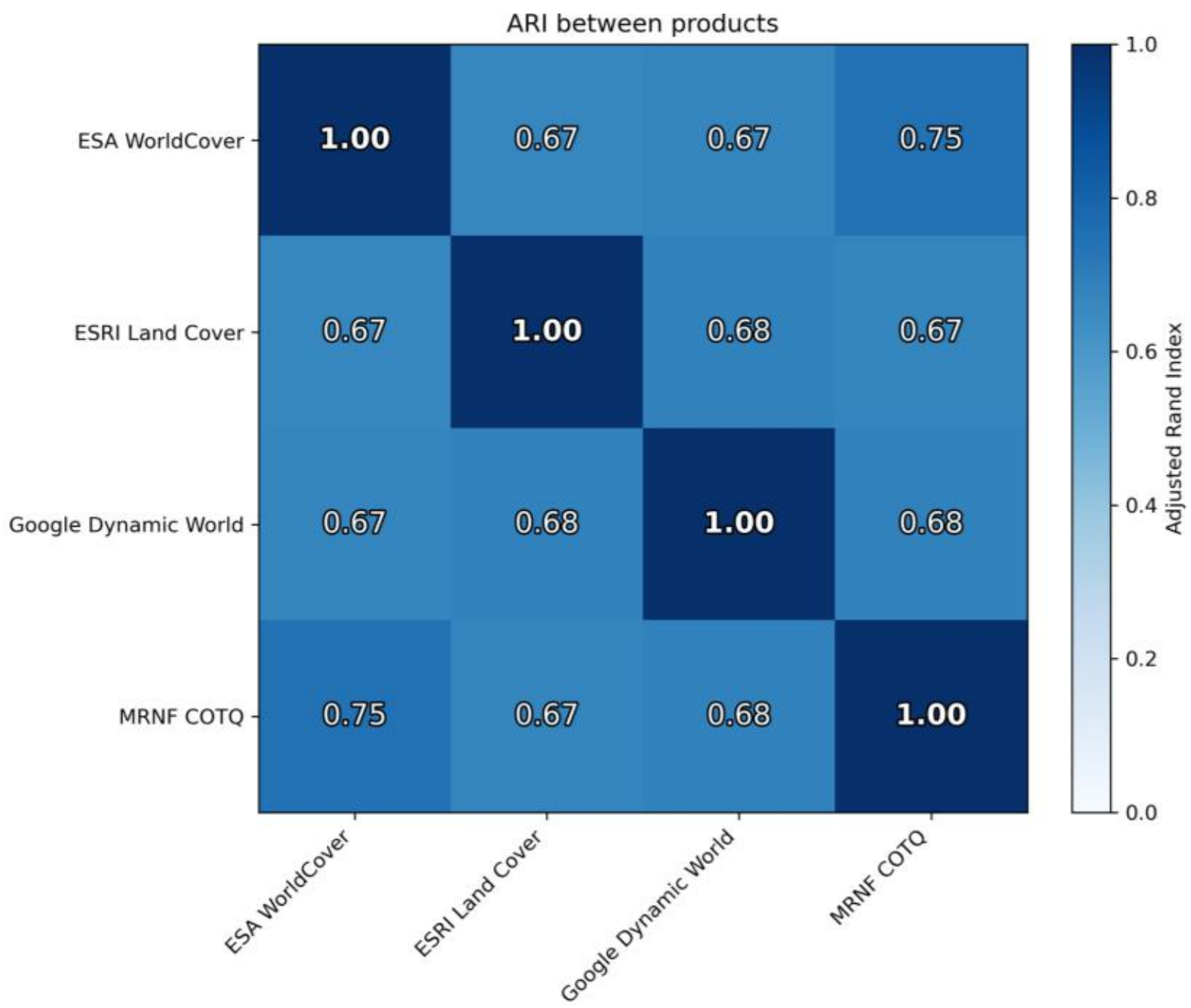

*Figure 13 : Pairwise Adjusted Rand Index (ARI) between the four land-cover products. The matrix reveals two product groupings: ESA and COTQ exhibit the highest mutual agreement (ARI = 0.75), while ESRI and Google show similar intermediate agreement levels with all other products.*

Importantly, this finding corroborates earlier observations derived from object-size distributions and shape complexities: both WorldCover and COTQ tend to preserve small structures and maintain more detailed boundary shapes compared to ESRI Land Cover and Google Dynamic World. The ARI therefore provides an independent confirmation that these two products segment the landscape in a more comparable manner.

The pairwise mean IoU matrix shown in Figure 14 reinforces this conclusion. IoU offers a more spatially explicit comparison by quantifying pixel-level agreement for each land-cover class before averaging across classes. The COTQ- WorldCover pair attains an IoU of 82 %, the highest among all pairs considered. In contrast, IoUs between COTQ and either Land Cover or Dynamic World remain lower (74 % -76 %).

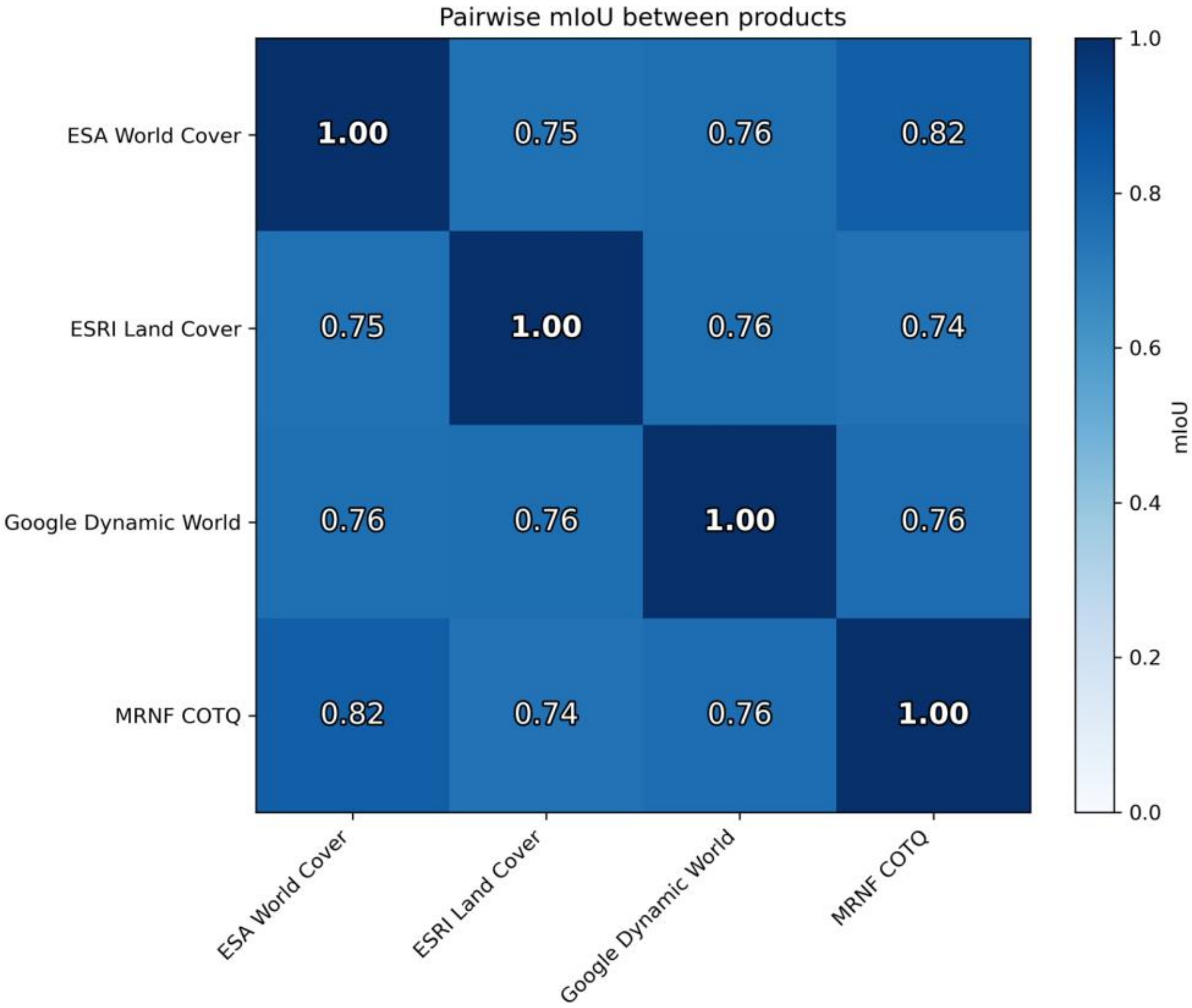

*Figure 14 : Pairwise mean Intersection-over-Union (mIoU) between the four land-cover products. As in the ARI matrix, the results reveal two groupings: ESA WorldCover and COTQ Land Cover show the highest mutual overlap (mIoU = 0.82), whereas ESRI Land Cover and Google Dynamic World exhibit more moderate and relatively uniform agreement levels with the other products*

#### *4.1.5 Selection of a reference product selection*

The object-size distribution, boundary-complexity, ARI, and mIoU analyses consistently identify ESA WorldCover as the global product whose spatial and structural behaviour is closest to that of COTQ. This convergence is meaningful because the four indicators capture complementary properties of the mapped landscape: ARI quantifies overall partition agreement, mIoU measures class-wise spatial overlap, and the object-size and boundary analyses characterize the preservation of fine-scale structures and geometric complexity. COTQ and WorldCover consistently form a distinct pair in these analyses, whereas ESRI Land Cover and Google Dynamic World exhibit substantially stronger spatial generalization, including fewer small objects and simpler object boundaries.

This result is also consistent with independent evidence from the literature. Recent comparative validation of the major global 10-m land-cover products identified WorldCover as the highest-performing product among WorldCover, Dynamic World, and ESRI Land Cover, confirming that it represents a particularly strong external benchmark rather than an arbitrary product selected because of its similarity to COTQ. The convergence between this independently established benchmark status and the structural and spatial results obtained here therefore places COTQ, at this first stage of the sequential evaluation, in the same overall performance tier as WorldCover and identifies the latter as the most informative global product against which COTQ should be examined more closely.

The subsequent photo-interpretation was therefore deliberately focused on COTQ / WorldCover disagreements. Its objective was not to repeat the initial four-product comparison using a second, redundant evaluation procedure, but to determine which of the two strongest products was more frequently supported by very-high-resolution reference imagery where their classifications genuinely differed. Extending the same interpretation campaign to ESRI Land Cover and Dynamic World would have substantially increased the validation effort while largely revisiting differences already characterized in the preceding structural and spatial analyses. WorldCover was therefore retained as the focused global comparator for the second stage of the evaluation.

## 4.2 Spectral separability metrics

In the absence of ground-truth data, the spectral separability of classes provides an indirect yet informative proxy for assessing the internal consistency and discriminative quality of each land-cover product. The following results are therefore interpreted as evidence of spectral organization, not as direct proof of thematic correctness. Three complementary indicators were used for this purpose: 1) class-wise Silhouette coefficients, 2) the global Calinski-Harabasz (CH) index, and 3) intra-class spectral angles derived from the Spectral Angle Mapper formulation. Spectral samples were extracted from Sentinel-2 RGB-NIR imagery acquired during the summer of 2021 and selected for minimal cloud contamination. Within each study area, the same image was used for all products to ensure that the comparison was performed under identical observation conditions. These metrics highlight how well each product partitions the four-band Sentinel-2 RGB-NIR feature space, acknowledging the inherent limitations associated with this relatively low-dimensional spectral domain. Indeed, this evaluation framework is constrained by the fact that it considers spectral information only, while many land-cover classes, particularly in heterogeneous environments, are distinguished not only by their reflectance values but also by spatial arrangement, and textural cues. Deep-learning models, unlike purely spectral metrics, fully exploit these spatial dependencies by analyzing neighbourhood context, edge patterns, object geometry, and textures. As a result, a product may exhibit modest spectral separability while still achieving strong discrimination through spatial-contextual reasoning (Blaschke, 2010; Zhu et al., 2017). Conversely, high spectral separability does not guarantee accurate mapping if spatial patterns are inconsistent or ambiguous. These considerations emphasize that spectral metrics, though useful, provide only a partial view of class distinctiveness compared to what is learned and leveraged by the underlying neural networks.

### *4.2.1 The Silhouette coefficient*

Figure 15 presents the Silhouette coefficient per class and per product. A positive Silhouette value indicates that a class is, on average, more similar to itself than to neighbouring classes, whereas negative values denote overlap or confusion in feature space.

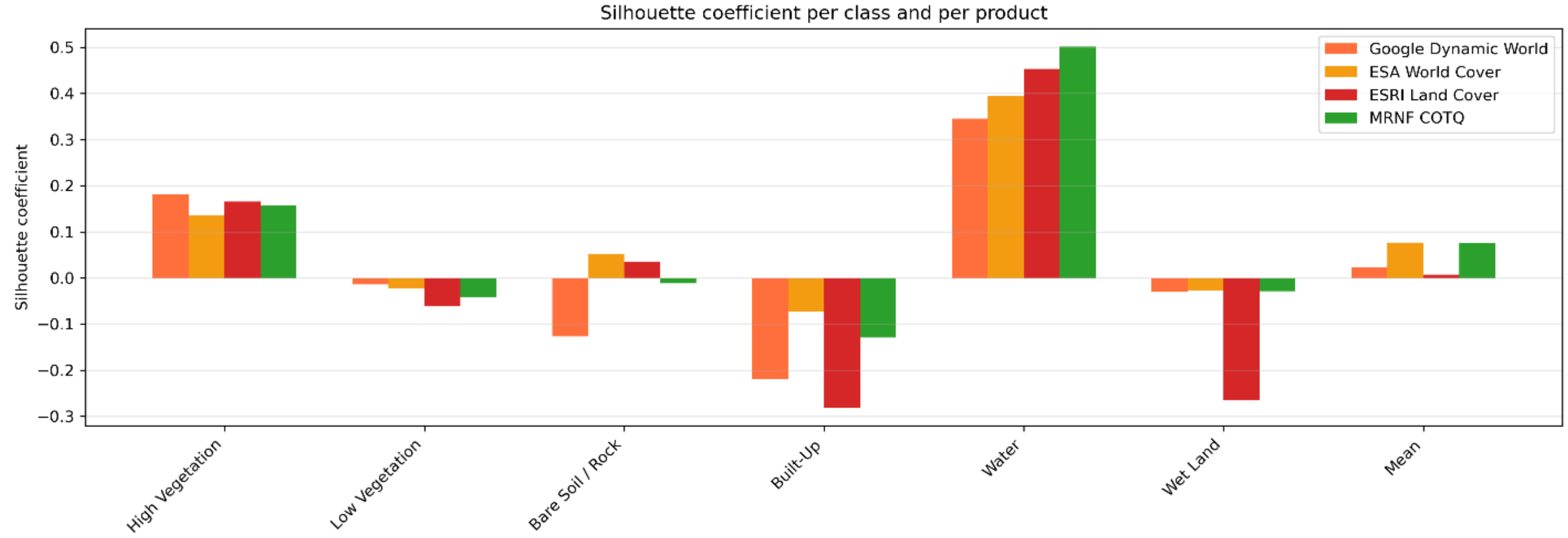

*Figure 15 : Silhouette coefficients computed per class for the four land-cover products. Water exhibits strong spectral separability for all products, whereas classes such as Built-Up, Low Vegetation, and Wetland show negative values, indicating substantial spectral overlap in the four-band RGB-NIR feature space.*

Across all products, Water exhibits the strongest separability, with Silhouette coefficients around 0.45-0.50, reflecting the unique spectral signature of water even in a reduced four-band configuration. High vegetation also shows modest positive separability for all products.

Low Vegetation, Bare Soil/Rock, Built-Up, and Wetland systematically yield negative Silhouette values. This pattern reflects the intrinsic challenge of separating these land-cover types using only four bands: their spectral responses often overlap, particularly in heterogeneous environments where vegetation, soil, and artificial surfaces coexist at the 10-m scale. Differences observed between products are therefore more indicative of class definitions or post-processing strategies than of absolute "quality".

Although the Sentinel-2 bands were standardized within each evaluation area before calculation, the Silhouette coefficient remains based on Euclidean distances between spectral vectors. It therefore remains sensitive to differences in spectral magnitude and relative band values within the standardized feature space, unlike SAM, which explicitly removes vector magnitude through L2 normalization and evaluates angular similarity. Consequently, illumination or surface-related variations that modify the relative position of pixels in the standardized feature space can still increase within-class dispersion and reduce Silhouette values.

The Silhouette analysis therefore suggests that all products face similar separability limitations in the reduced four-band RGB-NIR feature space, particularly for heterogeneous classes such as Low Vegetation, Built-Up, Rock/Bare, and Wetland. The SAM analysis presented in the following section provides a complementary assessment of spectral consistency by reducing sensitivity to spectral magnitude and focusing instead on spectral shape.

*4.2.3 Spectral Angle Mapper analysis*

SAM evaluates the angular difference between spectra and is therefore invariant to multiplicative illumination effects. Unlike the Silhouette coefficient, which is sensitive to variations in brightness or BRDF effects, SAM focuses exclusively on spectral shape, providing a more robust characterization of spectral consistency within each class.

Figure 16 presents the mean intra-class angle for each product.

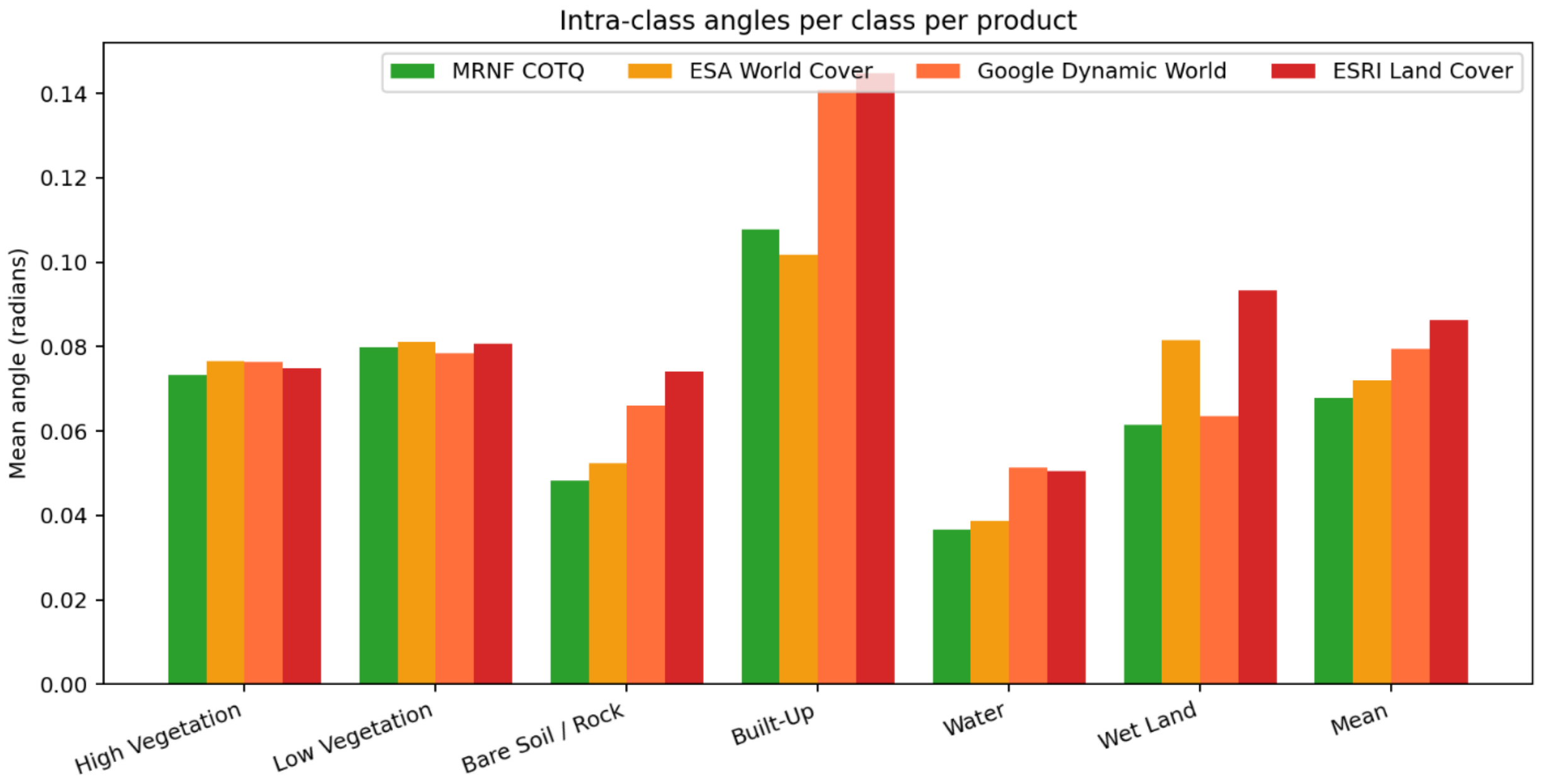

*Figure 16 : Mean intra-class spectral angles (in radians) computed using the SAM formulation for each land-cover class across the four products. Lower values indicate greater internal spectral consistency. COTQ and WorldCover generally exhibit the smallest intra-class angles.*

Across all products, Water, Bare Soil/Rock, and High Vegetation exhibit the smallest intra-class angles (approximately 0.035-0.040 radians, 0.050-0.070 radians, and 0.070-0.080 radians, respectively). This indicates that these classes maintain relatively consistent spectral shapes even within the reduced four-band RGB-NIR configuration.

For Water and High Vegetation, this behaviour aligns with their Silhouette score. In the case of Bare Soil/Rock, the improved intra-class consistency observed with SAM is likely linked to the metric's insensitivity to illumination intensity. Bare soils and rocky substrates often occur in rugged or sloped environments where reflectance magnitude can vary strongly with solar angle, terrain shading, and BRDF effects. Such illumination-driven variations would artificially inflate intra-class variability when using metrics based on absolute reflectance values. By comparing spectra through angular similarity rather than amplitude, SAM reduces this effect, thereby revealing a more intrinsic coherence of the class that was less apparent with the Silhouette coefficient.

Built-Up areas present the highest intra-class variability across all products (0.105-0.145 radians). This result is not surprising as built environments inherently mix multiple materials such as asphalt, concrete, roofing, bare soil patches, and vegetation. As a consequence, even the SAM metric, which is designed to be robust to illumination differences, reveals substantial internal heterogeneity for this class.

Wetland displays intermediate angles (0.065-0.095 radians), reflecting the diverse combinations of vegetation density, water content, and substrate characteristics typically found in wetland environments.

Although the class-wise patterns are consistent across all datasets, the products differ subtly in their internal consistency. COTQ Land Cover generally shows the lowest or among the lowest intra-class angles across most classes, suggesting slightly tighter spectral coherence. ESA WorldCover usually follows closely, with only marginally higher angles while Google Dynamic World tends to show slightly larger angles for several classes, particularly in Wet Land and Built-Up and ESRI Land Cover exhibits the highest intra-class variability overall, especially in Built-Up and Wet Land, where angles reach the upper end of the observed range.

These relative differences remain moderate, on the order of 0.005 to 0.015 radians, but they are consistent across classes, indicating that the COTQ and WorldCover products tend to maintain more compact spectral clusters than Google and ESRI.

#### *4.2.4 The Calinski-Harabasz index*

To complement the class-wise analyses produced by the Silhouette coefficients and the intra-class SAM angles, we computed the Calinski-Harabasz (CH) index, a global indicator of cluster separability that summarizes the balance between inter-class distance and intra-class compactness. Higher CH values indicate that class centroids are better separated relative to their internal dispersion, providing an integrated view of how consistently each product partitions the four-band Sentinel-2 reflectance space.

The CH scores reveal noticeable differences across the four land-cover products. It is presented in Table 4.

*Table 4: Calinski-Harabasz scores for each product.*

| | **Calinski-Harabasz scores** |
|---|---|
| **Google Dynamic World** | 3171 |
| **ESRI Land Cover** | 2871 |
| **ESA WorldCover** | 3479 |
| **COTQ** | 4351 |

COTQ Land Cover achieves the highest score (4351), followed by ESA WorldCover (3479), while Google Dynamic World (3171) and ESRI Land Cover (2871) exhibit lower values. Although the absolute magnitudes of these scores depend on class abundance and dataset size, their relative ordering is informative. The fact that COTQ and WorldCover obtain substantially higher CH values suggests that these products form more compact and better-separated clusters in spectral space. This trend is consistent with the results obtained using SAM angles, which showed lower intra-class variability for these two products, and with earlier observations based on object-size distributions and shape complexity, where COTQ and WorldCover also tended to preserve finer structural detail.

In contrast, the lower CH values of Land Cover and Dynamic World indicate broader intra-class dispersion or less distinct separation between classes. These results align with the previously noted tendency of these products to produce smoother classifications with fewer small objects and simpler shapes. Such smoothing or regularization may compromise cluster compactness when evaluated in a spectral feature space.

#### *4.2.5 Conclusion of the Spectral Separability Analysis*

Across all three separability indicators, Silhouette coefficients, SAM intra-class angles, and the global Calinski-Harabasz index, the results converge toward the same conclusion: COTQ (COTQ Land Cover) and ESA WorldCover exhibit the strongest spectral separability among the four evaluated products. Both datasets form more compact clusters in the four-band Sentinel-2 feature space and maintain clearer boundaries between classes. These findings are fully coherent with the SAM-based analysis, which showed systematically lower intra-class angles for COTQ and ESA, and with the CH index, where these two products achieved substantially higher scores than Google Dynamic World and ESRI Land Cover. Although the differences are moderate in absolute magnitude, the consistency of the pattern across all independent metrics indicates that COTQ and ESA WorldCover provide the most internally coherent spectral partitions, whereas Dynamic World and Land Cover tend to produce more diffuse or overlapping class distributions. This spectral evidence does not capture all aspects of classification quality, especially spatial and textural reasoning learned by deep networks, but it nonetheless reinforces the conclusion that COTQ and WorldCover possess a more robust spectral foundation on which their spatial classifications are built.

### 4.3 Photo-Interpretation

After identifying ESA WorldCover as the strongest global comparator, disagreement maps were computed between COTQ and WorldCover for each of the eight Sentinel-2 footprints. The photo-interpretation followed the protocol described in Section 3.6.2, including the possibility of COTQ-supported, WorldCover-supported, neither-product and indeterminate outcomes.

#### *4.3.1 The discrepancy maps*

The Figure 17 presents the spatial distribution of discrepancies between the WorldCover and COTQ products across a large landscape on the A evaluation site (Figure 7). Panel A shows all pixels where the two products disagree. As most of the discrepancies are border classes and fine roads or urban building already studied previously, we applied a three-pixels morphological erosion, highlighting the core discrepancies by removing boundary effects. For each evaluation site, 40 random sample points were then generated within the remaining disagreement areas and manually interpreted using very high-resolution imagery, resulting in a total of 320 photo-interpreted observations across the eight study areas. Panel B shows the disagreement areas remaining after the three-pixel morphological erosion, while Panel C shows the spatial distribution of the photo-interpretation sample points drawn from these same eroded disagreement areas, providing spatial coverage of the retained substantive disagreements across the evaluation site. These figures show that, although disagreements are mainly concentrated in ecologically complex areas such as urban fringes and wetlands, the two products remain largely consistent, exhibiting strong structural agreement across extensive homogeneous regions.

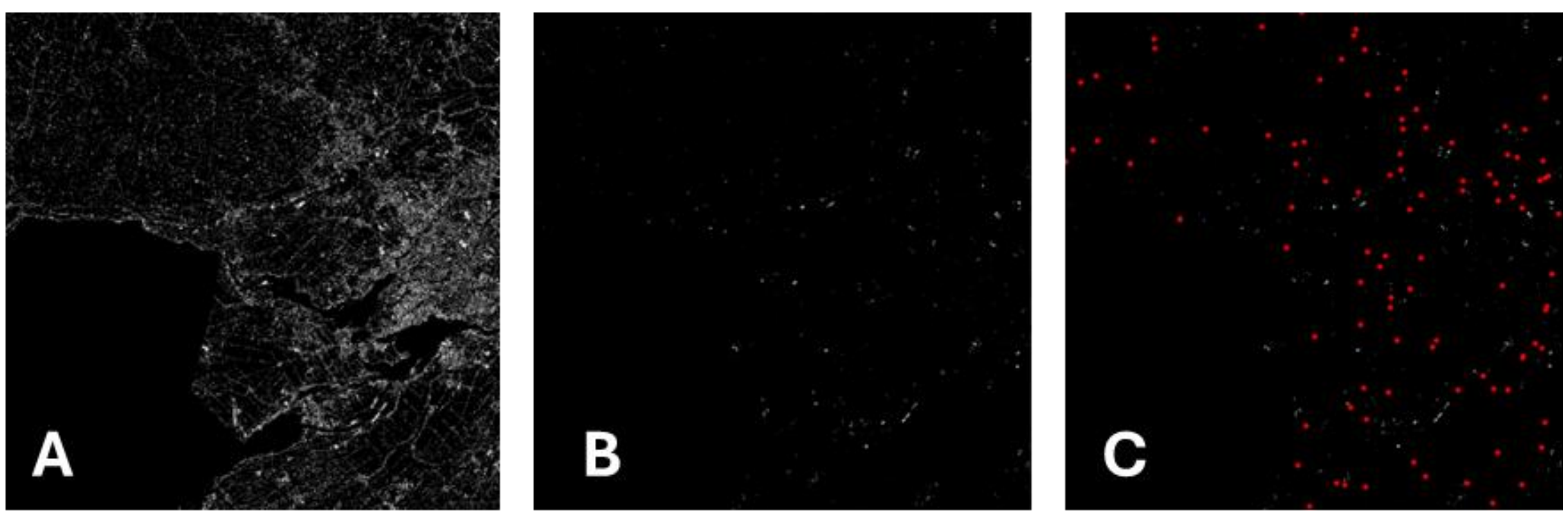

*Figure 17 : Visualization of disagreement locations between the COTQ and WorldCover products. (A) Raw disagreement map. (B) Disagreement map after a 3-pixel morphological erosion to remove boundary effects. (C) Distribution of the points selected for manual photo-interpretation.*

*4.3.2 Photo-interpretation outcomes within disagreement areas*

The figure 18 presents the count of photo-interpreted points correctly classified by each product per class. Across the eight evaluation sites, 320 disagreement points were photo-interpreted. The resulting outcomes are summarized in Table 5. Indeterminate observations were excluded from the quantitative COTQ-WorldCover comparison because no sufficiently confident reference interpretation could be established, whereas cases in which neither product matched the reference interpretation were retained as a separate outcome.

*Table 5: Overall photo-interpretation outcomes for the 320 sampled disagreement points.*

| Photo-interpretation outcome | Number of points | Included in class-wise COTQ–WorldCover comparison |
|---|---|---|
| **COTQ supported** | 162 | Yes |
| **WorldCover supported** | 75 | Yes |
| **Neither product supported** | 4 | No |
| **Indeterminate** | 79 | No |
| **Total** | 320 | - |

Among the interpretable sampled disagreements, COTQ was supported more frequently for Low Vegetation, Rock/Bare Soil, Built-Up, Water, and Wetland, whereas WorldCover was more frequently supported for Forest. Overall, the class-wise outcomes within the sampled disagreement domain favoured COTQ for most of the evaluated classes. However, these results describe relative thematic support within the sampled disagreement areas and should not be interpreted as class-specific accuracy estimates over the complete study domain.

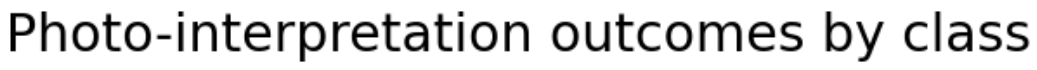

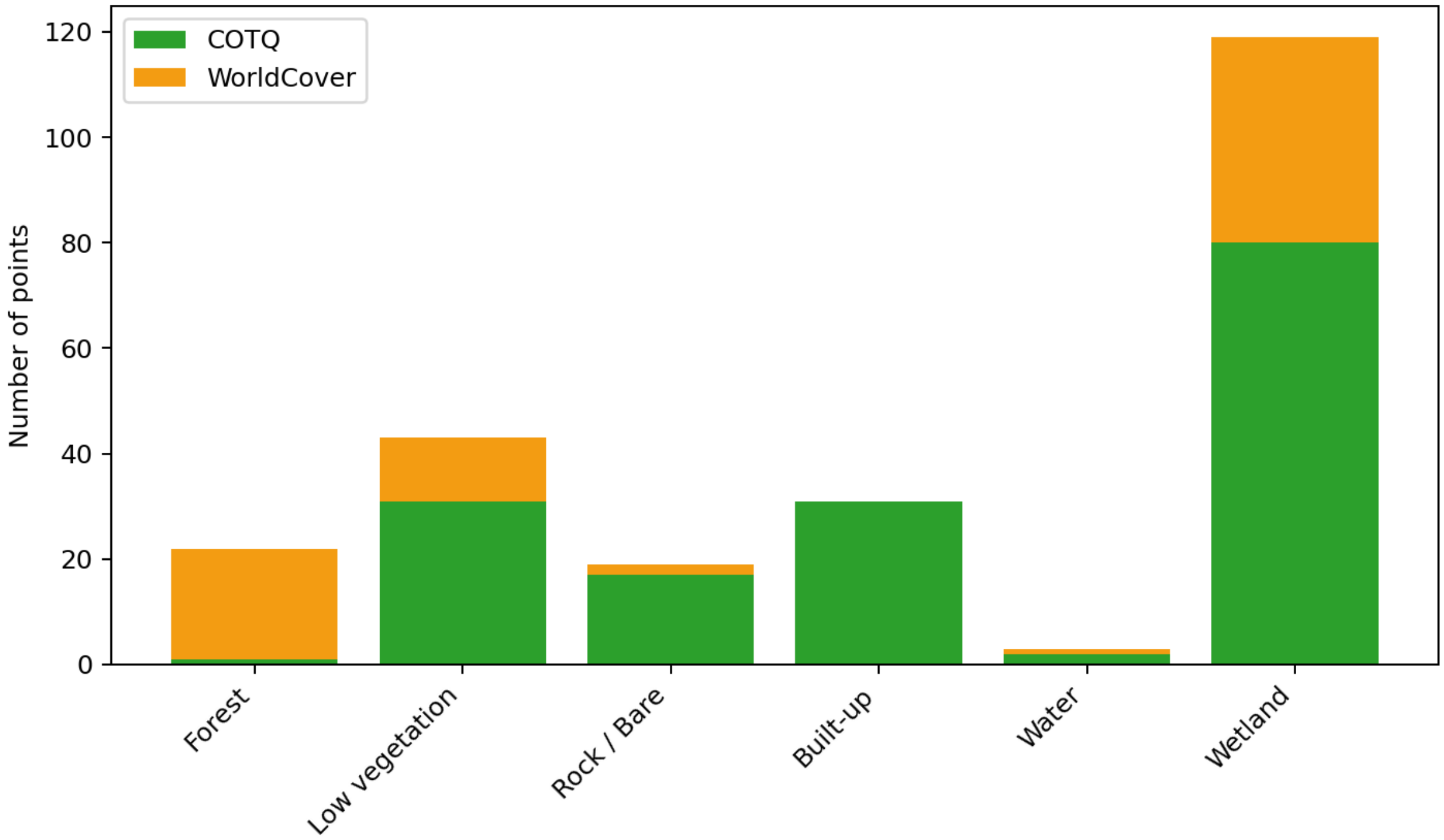

*Figure 18 : Photo-interpretation outcomes by land-cover class within the sampled substantive disagreement areas between COTQ and WorldCover. Bars show the number of observations for which very-high-resolution reference imagery supported either COTQ or WorldCover. Only observations for which one of the two products could be confidently identified as correct are included; indeterminate observations and cases in which neither product was judged correct are excluded from this class-based comparison.*

*4.3.3 Example-driven assessment of product disagreements and interpretation limitations*

The following figures, each showing very high-resolution imagery, ESA classification, and COTQ classification, offer a closer look at how the two products differ in specific environments.

*4.3.3.1 Urban and anthropogenic surfaces*

In the first example (Figure 19, A,B and C), WorldCover clearly misclassifies a large asphalt parking area as Low Vegetation, presumably because scattered shrubs and patches of grass are visible around and within the lot. Although vegetation is present, the underlying surface is unequivocally artificial and possibly impervious, and the area functions entirely as built infrastructure. In this case, the COTQ classification of Built-Up is unambiguously correct, and the ESA error reflects an alredy shown tendency of the WorldCover product to assign mixed urban pixels containing even small amounts of vegetation to vegetation classes, thereby underestimating urbanized surfaces in heterogeneous urban mosaics.

The second example (Figure 19, D, E and F), however, illustrates a different situation. Here, WorldCover classifies large earthworks, quarries, or construction sites as Bare Soil / Rock, whereas COTQ labels them as Built-Up. Unlike the parking lot, these cases are genuinely ambiguous: physically, the exposed substrate is indeed soil or rock, yet from a land-use perspective these areas correspond to active anthropogenic modification of the landscape. Both classifications are therefore defensible depending on whether the focus is placed on physical land cover (WorldCover) or functional soil artificialization (COTQ). This distinction underscores that part of the disagreement between products does not originate from misclassification, but rather from legitimate differences in conceptual definitions of what constitutes "urban" or "artificial" land.

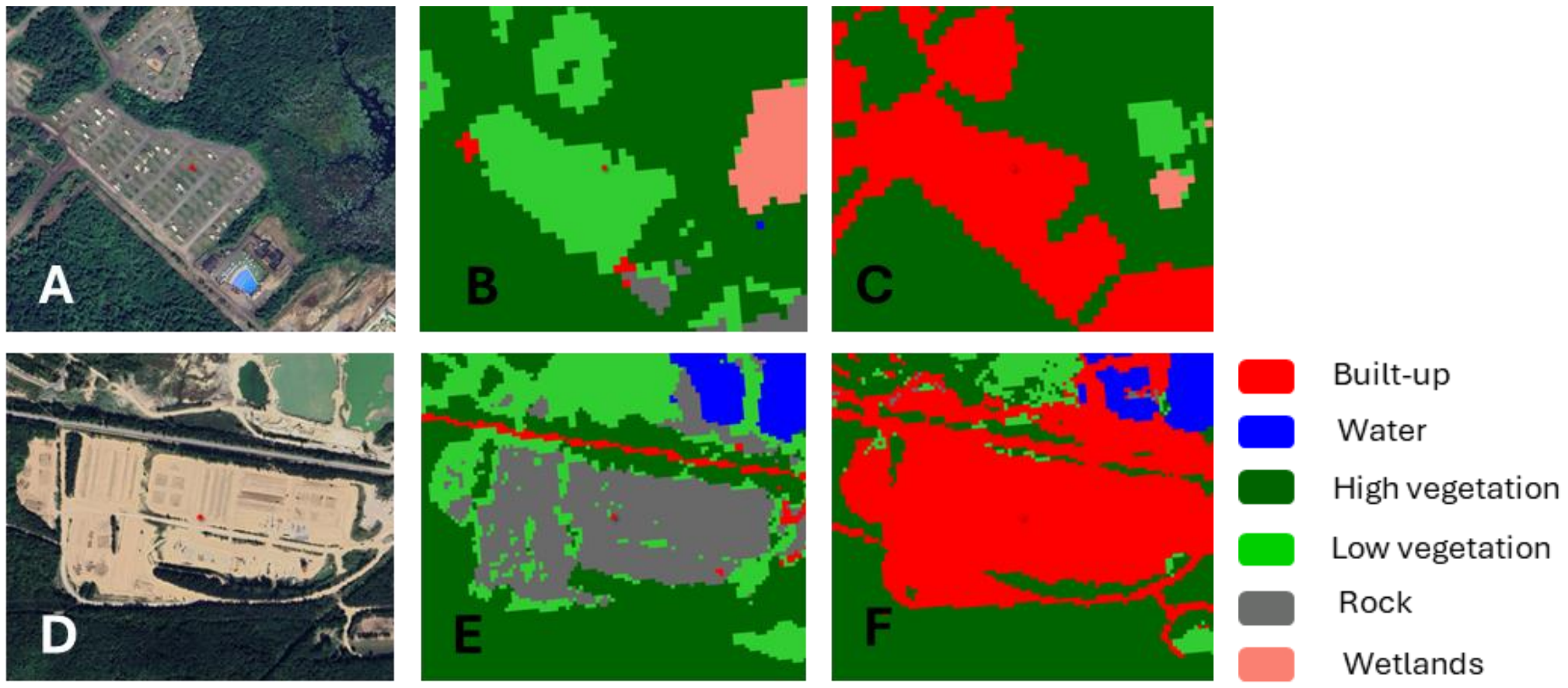

*Figure 19 : (A) Very high-resolution reference imagery. (B) ESA WorldCover classification. (C) COTQ classification. In this first example, ESA incorrectly assigns a large asphalt parking lot to Low Vegetation, whereas COTQ correctly maps the area as Built-Up. (D) Very high-resolution reference imagery of an active earthwork zone. (E) ESA WorldCover classification, which maps most of the site as Bare Soil / Rock, reflecting the visible physical substrate. (F) COTQ classification, which labels these surfaces as Built-Up, consistent with its definition of artificialized land.*

This distinction is especially important because one of the major practical uses of an annual repeated land-cover product is to monitor changes in land use and soil artificialization over time. From this perspective, the COTQ considered it essential that the emergence of new quarries, mining pits, large earthworks, or major construction zones be visible as a clear transition toward artificial land. If these human-induced transformations were systematically labeled as Bare Soil / Rock, as in the ESA product, major developments such as the opening of a mine, which profoundly alters the landscape, could remain invisible in long-term artificialization analyses simply because the exposed substrate is mineral in nature. In this sense, the COTQ classification approach prioritizes functional change detection over purely physical cover, which is why these areas are grouped under Built-Up despite their rocky appearance.

#### *4.3.3.2 Wetlands: peatlands vs. riparian wetlands*

The two examples in Figure 20 highlight how different wetland subclasses challenge each product in distinct ways, consistent with the diversity of wetland types defined in the MELCCFP system.

In the first example, the very high-resolution imagery shows a lake shoreline wetland composed of saturated soils, emergent vegetation, and irregular patches of hydrophytic plants, a configuration typical of shallow-water marshes and riparian wetlands. These systems exhibit fine-scale intermixing between water, low vegetation, and adjacent forest cover, forming narrow and highly fragmented ecological gradients. In this case, WorldCover correctly preserves most of the shoreline wetland, while COTQ tends to incorrectly label substantial portions of it as Forest.

In contrast, the second example shows a peatland / wet meadow complex, characterized by an open physiognomy, dense graminoids or mosses, and little to no forest canopy, a configuration typical of bogs, fens, and grass-dominated wet meadows. These ecosystems form large, spatially coherent patches with distinctive vegetation structure. In this example, the COTQ product correctly maps the entire peatland-wet meadow complex as a continuous wetland unit, whereas WorldCover underestimates its extent by fragmenting the wetland and reassigning substantial portions of the peatland to the Low Vegetation class. From a strictly technical perspective, this is not incorrect: peat-forming vegetation such as sphagnum mosses, sedges, and dwarf shrubs does fall within the definition of low vegetation. However, labeling these surfaces as Low Vegetation fails to capture the true ecological extent of the wetland. Hence, the ESA classification appears to break the peatland into smaller patches, while the COTQ product more accurately preserves the continuity of the wetland complex.

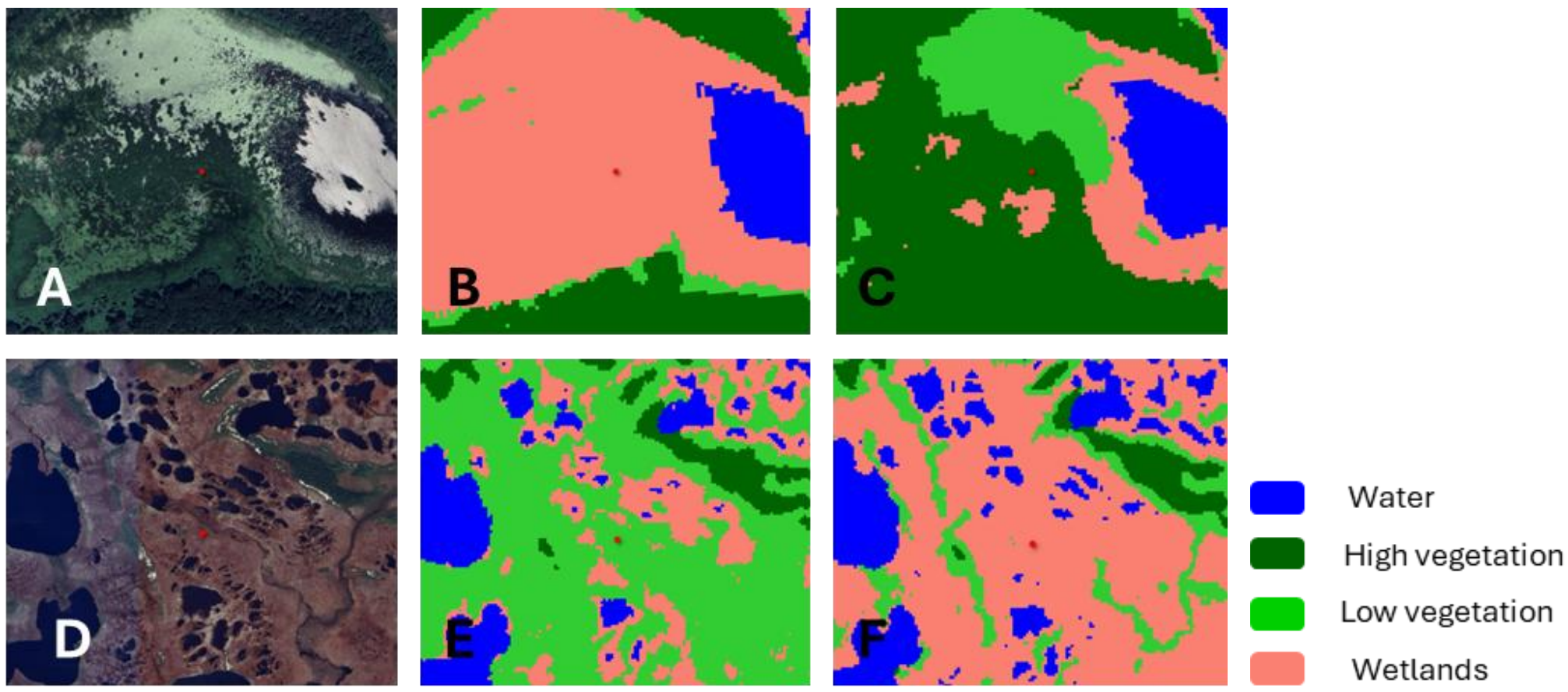

*Figure 20 : (A-C) Example of a shoreline wetland along a lake margin. A: Very high-resolution reference imagery.*
*B: ESA WorldCover classification, correctly capturing the extent of the shoreline wetland. C: COTQ Land Cover classification, underestimating the wetland and mapping parts of it as Forest. (D-F) Example of a peatland / wet meadow complex. D: Very high-resolution reference imagery. E: ESA WorldCover classification, fragmenting the peatland and assigning large areas to Low Vegetation. F: COTQ Land Cover classification, correctly delineating the full extent of the peatland.*

Together, these two examples demonstrate that neither product is uniformly superior across all wetland subclasses. WorldCover tends to perform better in riparian and shoreline wetlands, where fine-scale hydrological gradients dominate, while COTQ is better in peatlands and wet meadows, which present more coherent structural and spectral signatures.

*4.3.3.3 Moss, lichen, and low vegetation over bedrock*

In northern environments, the very high-resolution imagery reveals mosaics of exposed bedrock partially covered by mosses and lichens, an ecologically important vegetation type that occurs extensively across the northern regions of Québec. Although the ESA WorldCover product includes a dedicated “Moss & Lichen” class, neither Google Dynamic World nor ESRI Land Cover provides an equivalent category. To ensure comparability across products, these surfaces had to be aggregated into a broader Low Vegetation class during the post-processing stage. As a result, all sparse, and cryptogamic vegetation types of the northern shield were grouped under Low Vegetation for evaluation purposes. The “errors” presented under this label in the following examples therefore refer specifically to moss and lichen dominated rocky environments, not to low vegetation in the general sense.

Nearly all disagreements between WorldCover and COTQ in these northern regions correspond to rocky surfaces partially covered by mosses and lichens. WorldCover maps extensive areas using its dedicated

“Moss & Lichen” class, whereas COTQ assigns the same surfaces predominantly to Rock/Bare (Figures 21 and 22).

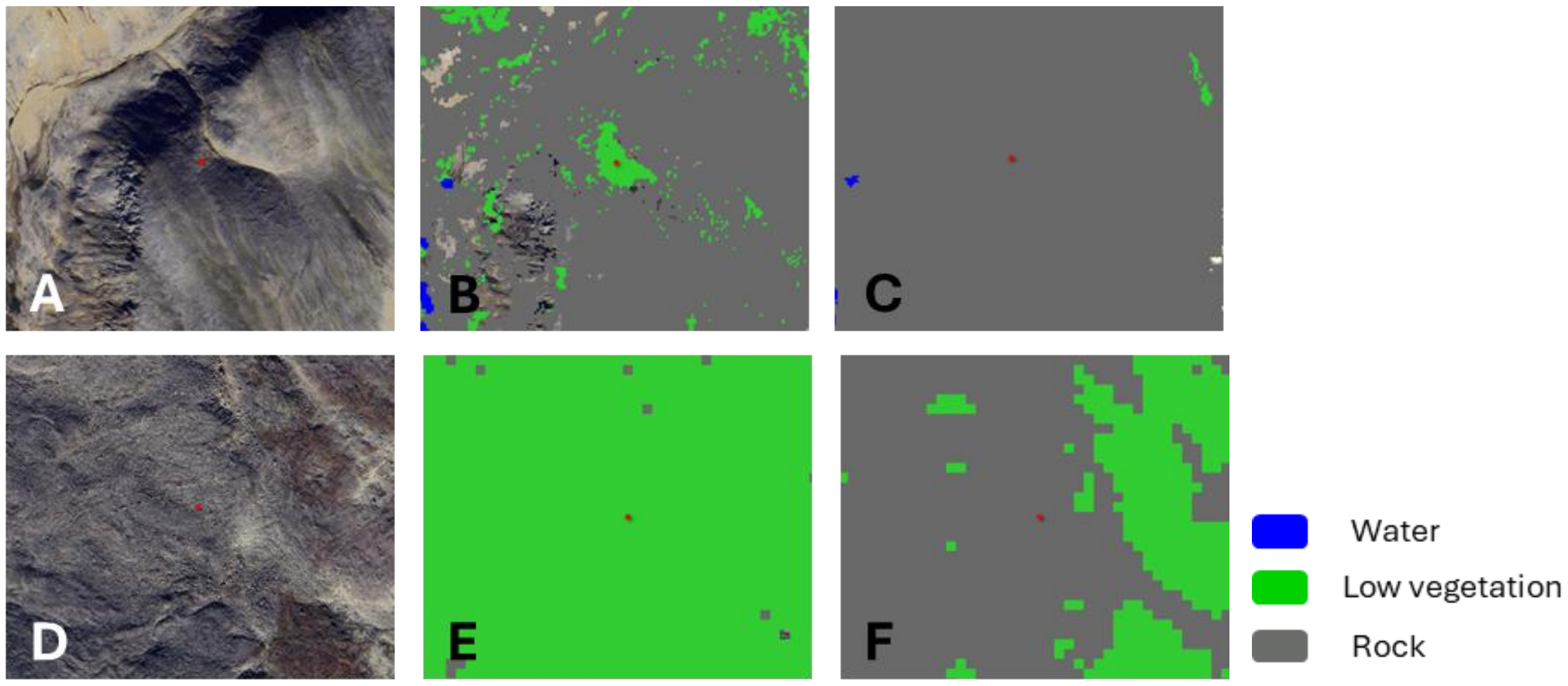

*Figure 21 : Examples of WorldCover-COTQ discrepancies in northern rocky environments at the point level. (A, D) Very high-resolution reference imagery. (B, E) ESA WorldCover classifications. (C, F) COTQ Land Cover classifications. Panels illustrate two instances of the same systematic disagreement between the two products. In both examples, ESA assigns large portions of the rocky landscape to Low Vegetation (interpreted as moss and lichen cover), whereas COTQ classifies nearly all of these surfaces as Bare Rock.*

Interpreting these systematic disagreements requires consideration of the spatial support of the products being evaluated. At 10-m resolution, a single thematic label must represent a 100-m² surface, even though several land-cover components may coexist within that area. The objective of the reference interpretation is therefore not to determine whether mosses or lichens are completely absent from a pixel, but to identify the land-cover class that is best supported at the spatial scale of the mapped product. Where very-high-resolution imagery clearly shows exposed bedrock or regolith as the dominant observable surface, while cryptogamic vegetation cannot be visually confirmed even at a substantially finer spatial resolution, assigning the entire 10-m pixel to Low Vegetation or Moss & Lichen would not constitute a better-supported interpretation. In such cases, when COTQ assigns Rock/Bare and WorldCover assigns Low Vegetation, the available reference evidence supports the COTQ interpretation. This does not imply that cryptogamic vegetation is biologically absent from the pixel; rather, it indicates that Rock/Bare is the thematic class most strongly supported by the available reference evidence at the 10-m mapping scale. Observations for which neither exposed rock nor cryptogamic vegetation could be identified with sufficient confidence as the appropriate pixel-level interpretation were recorded as indeterminate and excluded from the quantitative comparison.

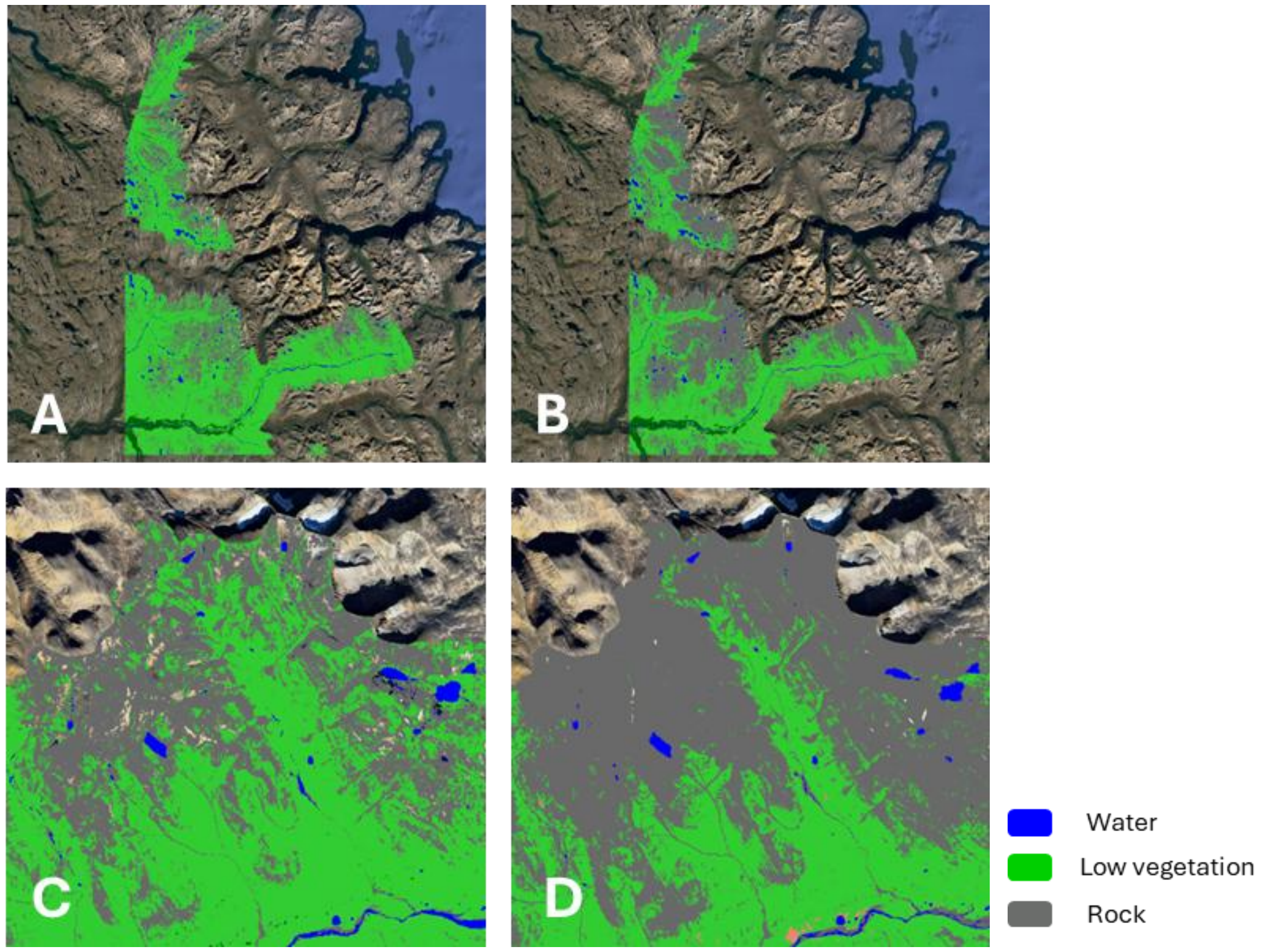

*Figure 22 : Large-scale comparison of WorldCover (A, C) and COTQ (B, D) classifications in northern Québec. (A, B) Regional view displaying the spatial extent of Low Vegetation (green) and Bare Rock (gray) classes for WorldCover (A) and COTQ (B). (C, D) Zoomed-in view showing the detailed spatial patterns of disagreement at the same location. ESA maps extensive areas as Low Vegetation (interpreted as Moss & Lichen), whereas COTQ classifies most rocky surfaces as Bare Rock.*

## 5 Conclusion

This study provides the first comprehensive evaluation of the Cartographie de l'Occupation du Territoire Québécois (COTQ), a 10-m annual land-cover product developed for the Ministère des Ressources naturelles et des Forêts for province-wide monitoring of land use and soil artificialisation. The COTQ mapping was produced using a semi-supervised learning strategy based on large-scale pseudo-synthetic training data generated through automated pre-segmentation and selective polygon curation, substantially reducing the need for exhaustive manual vector annotation at 10-m resolution. Unlike previous efforts focused on describing the production pipeline, the present work concentrated on objectively assessing the behaviour, internal consistency, and comparative performance of the COTQ relative to the three major global 10-m LULC datasets, ESA WorldCover, Google Dynamic World, and ESRI Land Cover, using a suite of structural, spectral, and expert-based validation approaches.

Across all structural indicators, object-size distributions, boundary complexity, Adjusted Rand Index, and pairwise IoU, the analyses consistently showed that COTQ and ESA WorldCover form a distinct pair with markedly higher spatial detail and finer segmentation internal consistency than the ESRI and Google products. Both COTQ and WorldCover preserve small objects, maintain complex boundary geometries, and exhibit stronger Zipf-like object-size patterns than ESRI Land Cover and Google Dynamic World. These properties are essential for applications that rely on detecting subtle changes in anthropogenic surfaces, narrow wetlands, or sparse northern vegetation. The converging structural, spatial, and spectral analyses identified WorldCover as the strongest and most informative global comparator among the evaluated products, justifying its selection for the subsequent disagreement and photo-interpretation analyses.

Spectral separability metrics, Silhouette scores, SAM intra-class angles, and Calinski-Harabasz indices, converged toward the same conclusion. Among the four products, COTQ and WorldCover consistently exhibited the most compact and best-separated spectral clusters in Sentinel-2 RGB-NIR space. Although these metrics provide only a partial view of class distinctiveness, they nonetheless demonstrate that COTQ and WorldCover are grounded in more coherent spectral partitions than the more generalized ESRI and Google products. These spectral findings are consistent with the structural results and support the interpretation that the additional spatial detail preserved by COTQ is not solely attributable to noise. However, because the spectral analysis is restricted to four Sentinel-2 bands and does not account for spatial context or temporal information, it should be interpreted as a complementary diagnostic indicator rather than as direct evidence of classification accuracy.

The expert photo-interpretation assessment further clarified the nature of disagreements between ESA WorldCover and the COTQ. In urban mosaics and anthropogenic surfaces, WorldCover tended to underestimate built-up areas where vegetation is intermixed, whereas COTQ more faithfully captured the functional extent of artificial surfaces, an essential requirement for monitoring soil artificialisation. While this interpretation remains subject to expert judgement and class-definition choices, it provides useful insight into the practical implications of the observed disagreements. Conversely, in some riparian wetlands, WorldCover better preserved narrow hydrological and vegetation gradients that COTQ occasionally assigned to Forest, illustrating that WorldCover can outperform COTQ in specific ecological contexts. In northern environments, the two products diverged primarily in their representation of rocky surfaces partially covered by cryptogamic vegetation. Because the reference interpretation was conducted at the 10-m mapping scale, cases where exposed rock or regolith was the dominant observable surface were supported as Rock/Bare, whereas observations that could not be interpreted with sufficient confidence were recorded as indeterminate and excluded from the quantitative comparison. These example-driven analyses underscore that differences between products do not uniformly reflect classification errors but often arise from distinct thematic definitions and domain-specific priorities. However, despite its strong relative performance, the operational usefulness of ESA WorldCover for long-term monitoring in Québec is fundamentally constrained by the absence of publicly available updates beyond the 2021 release. As a result, even though WorldCover rivals COTQ in several environments, its lack of temporal continuity prevents it from supporting annual change-detection applications critical to provincial needs.

All evaluation angles, structural, spectral, and interpretative, provide converging evidence that the COTQ product offers a robust, detailed, and thematically coherent representation of Québec's land cover. Under the proposed evaluation framework, its performance is consistently comparable to, and in several cases more favourable than, that of the global products considered in this study. Crucially, COTQ preserves fine-scale anthropogenic structures and offers stable annual updates based on classification of every Sentinel-2 acquisition rather than pre-composited imagery. This full temporal transparency, combined with the capacity to adapt class definitions to provincial needs, provides a decisive operational advantage for monitoring soil artificialisation, land-use transitions, and infrastructure development across Québec.

However, several limitations should be acknowledged. Spectral assessments were restricted to four bands, and the lack of province-scale ground truth remains a fundamental constraint. The photo-interpretation campaign, although systematic and stratified, is inherently influenced by the limitations of available high-resolution imagery, particularly in northern cryptogamic environments. Many of the disagreements observed throughout this study should not be interpreted solely as classification errors. In several cases, they reflect different thematic objectives embedded in the design of the products themselves. Global products such as ESA WorldCover, ESRI Land Cover, and Dynamic World primarily aim to provide globally consistent representations of physical land cover, whereas COTQ was specifically developed to support land management, artificialisation monitoring, and territorial planning within Québec. Consequently, some differences arise from alternative but internally consistent interpretations of the same landscape features rather than from unequivocal classification errors.

Despite these limitations and differences in thematic objectives, the complementary analyses converge toward a consistent overall assessment of the relative performance of the four products, summarized in Table 6.

*Table 6: Summary of the main comparative results across the four evaluated land-cover products.*

| **Evaluation component** | **COTQ** | **ESA WorldCover** | **Google Dynamic World** | **ESRI Land Cover** | **Main interpretation** |
|---|---|---|---|---|---|
| Object-size distribution | High fine-scale detail | High fine-scale detail | More generalized | More generalized | COTQ and WorldCover preserve substantially more small objects |
| Boundary complexity | High | High | Lower | Lower | COTQ and WorldCover preserve more intricate geometries |
| Agreement with COTQ | - | Highest: ARI = 0.75; mIoU = 0.82 | Lower | Lower | WorldCover is the closest global comparator |
| Built-Up IoU vs. 50-cm reference | **67.5%** | 52.5% | 57.4% | 55.2% | COTQ shows the strongest agreement with the localized high-resolution Built-Up reference |
| Spectral separability | Strongest overall | Strong | Intermediate | Lowest overall | COTQ and WorldCover provide the most coherent four-band spectral partitions |
| Calinski–Harabasz | **4351** | 3479 | 3171 | 2871 | Same overall spectral ordering |
| Photo-interpretation of disagreements | **162 supported** | 75 supported | Not assessed | Not assessed | Among interpretable COTQ–WorldCover disagreements, COTQ was supported more frequently |
| Operational temporal capability | Annual COTQ production | No public update beyond 2021 | - | - | COTQ better matches the annual provincial monitoring objective |

Beyond the specific case of Québec, the proposed evaluation framework may also be relevant for other regional, provincial, or national land-cover products. In situations where no single authoritative reference dataset is available, the combination of semantic harmonization, structural indicators, spectral analyses, and targeted photo-interpretation provides a practical means of comparing competing products while explicitly acknowledging their respective limitations. Although several aspects of the present study are tied to Québec's ecological conditions and operational priorities, the broader methodological framework remains transferable to other jurisdictions facing similar challenges related to reference data availability, class-definition differences, and product interoperability.

Paraphrasing Box and Draper (1987), all land-cover maps are necessarily imperfect representations of reality, but some are more useful than others for a given purpose. The results of this study show that the development of COTQ was justified precisely in this operational sense. Within the Québec environmental conditions, harmonized classes, and monitoring objectives evaluated here, COTQ provides the strongest overall combination of spatial detail, thematic suitability, comparative support, and temporal update capability among the four products considered. For users requiring a 10-m land-cover dataset specifically adapted to Québec and capable of supporting annual monitoring of land occupation and soil artificialisation, COTQ therefore represents the preferred operational product among the evaluated alternatives. ESA WorldCover remains a strong global benchmark, but its globally standardized taxonomy and absence of publicly available updates beyond 2021 limit its suitability for several provincial applications. The value of COTQ does not lie in claiming to provide a perfect representation of the territory, but in providing one that is demonstrably better adapted to the questions Québec users need to answer.

## 6 References

Adedeji, O., Owoade, P., Ajayi, O., and Arowolo, O. 2022. "Image Augmentation for Satellite Images." doi:10.48550/arXiv.2207.14580.

Benhammou, Y., Alcaraz-Segura, D., Guirado, E., Khaldi, R., Achchab, B., Herrera, F., and Tabik, S. 2022. "Sentinel2GlobalLULC: A Sentinel-2 RGB image tile dataset for global land use/cover mapping with deep learning." *Scientific Data*, Vol. 9 (No. 1), p. 681. doi:10.1038/s41597-022-01775-8.

Blaschke, T. 2010. "Object based image analysis for remote sensing." *ISPRS Journal of Photogrammetry and Remote Sensing*, Vol. 65 (No. 1), pp. 2–16. doi:10.1016/j.isprsjprs.2009.06.004.

Box, G.E.P., and Draper, N.R. 1987. *Empirical Model-Building and Response Surfaces*. Wiley Series in Probability and Mathematical Statistics. John Wiley & Sons.

Brown, C.F., Brumby, S.P., Guzder-Williams, B., Birch, T., Hyde, S.B., Mazzariello, J., Czerwinski, W., Pasquarella, V.J., Haertel, R., Ilyushchenko, S., Schwehr, K., Weisse, M., Stolle, F., Hanson, C., Guinan, O., Moore, R., and Tait, A.M. 2022. "Dynamic World, Near real-time global 10 m land use land cover mapping." *Scientific Data*, Vol. 9 (No. 1), p. 251. doi:10.1038/s41597-022-01307-4.

Calinski, T., and Harabasz, J. 1974. "A dendrite method for cluster analysis." *Communications in Statistics - Theory and Methods*, Vol. 3 (No. 1), pp. 1–27. doi:10.1080/03610927408827101.

Chen, L.-C., Papandreou, G., Schroff, F., and Adam, H. 2017. "Rethinking Atrous Convolution for Semantic Image Segmentation." doi:10.48550/arXiv.1706.05587.

Clabaut, É., Foucher, S., Bouroubi, Y., and Germain, M. 2024. "Synthetic Data for Sentinel-2 Semantic Segmentation." *Remote Sensing*, Vol. 16 (No. 5), p. 818. doi:10.3390/rs16050818.

*Classification écologique du territoire québécois*. 2021. Ministère des forêts, de la faune et des parcs.

Felzenszwalb, P.F., and Huttenlocher, D.P. 2004. "Efficient Graph-Based Image Segmentation." *International Journal of Computer Vision*, Vol. 59 (No. 2), pp. 167–181. doi:10.1023/B:VISI.0000022288.19776.77.

Jaccard, P. 1901. "Étude comparative de la distribution florale dans une portion des Alpes et du Jura." doi:10.5169/SEALS-266450.

Karra, K., Kontgis, C., Statman-Weil, Z., Mazzariello, J.C., Mathis, M., and Brumby, S.P. 2021. "Global land use / land cover with Sentinel 2 and deep learning." 2021 IEEE International Geoscience and Remote Sensing Symposium IGARSS, pp. 4704–4707. Presented at the IGARSS 2021 - 2021 IEEE International Geoscience and Remote Sensing Symposium, IEEE. doi:10.1109/IGARSS47720.2021.9553499.

Kruse, F.A., Heidebrecht, K.B., Shapiro, A.T., Barloon, P.J., and Goetz, A.F.H. n.d. “The Spectral Image Processing System (SIPS) Interactive Visualization and Analysis of Imaging Spectrometer Data.”

Lei, C., Hu, B., Wang, D., Zhang, S., and Chen, Z. 2019. “A Preliminary Study on Data Augmentation of Deep Learning for Image Classification.” Proceedings of the 11th Asia-Pacific Symposium on Internetware, pp. 1–6. Presented at the Internetware ’19: The 11th Asia-Pacific Symposium on Internetware, ACM. doi:10.1145/3361242.3361259.

Lozano-Tello, A., Siesto, G., Fernández-Sellers, M., and Caballero-Mancera, A. 2023. “Evaluation of the Use of the 12 Bands vs. NDVI from Sentinel-2 Images for Crop Identification.” *Sensors*, Vol. 23 (No. 16), p. 7132. doi:10.3390/s23167132.

Malinowski, R., Lewiński, S., Rybicki, M., Gromny, E., Jenerowicz, M., Krupiński, Michał, Nowakowski, A., Wojtkowski, C., Krupiński, Marcin, Krätzschmar, E., and Schauer, P. 2020. “Automated Production of a Land Cover/Use Map of Europe Based on Sentinel-2 Imagery.” *Remote Sensing*, Vol. 12 (No. 21), p. 3523. doi:10.3390/rs12213523.

Masoudi, M., Richards, D.R., and Tan, P.Y. 2024. “Assessment of the Influence of Spatial Scale and Type of Land Cover on Urban Landscape Pattern Analysis Using Landscape Metrics.” *Journal of Geovisualization and Spatial Analysis*, Vol. 8 (No. 1), p. 8. doi:10.1007/s41651-024-00170-8.

Moraes, D., Campagnolo, M.L., and Caetano, M. 2024. “Training data in satellite image classification for land cover mapping: a review.” *European Journal of Remote Sensing*, Vol. 57 (No. 1), p. 2341414. doi:10.1080/22797254.2024.2341414.

Phiri, D., Simwanda, M., Salekin, S., Nyirenda, V., Murayama, Y., and Ranagalage, M. 2020. “Sentinel-2 Data for Land Cover/Use Mapping: A Review.” *Remote Sensing*, Vol. 12 (No. 14), p. 2291. doi:10.3390/rs12142291.

Rad, R. 2024. “Vision Transformer for Multispectral Satellite Imagery: Advancing Landcover Classification*.” 2024 IEEE/CVF Winter Conference on Applications of Computer Vision (WACV), pp. 8161–8168. Presented at the 2024 IEEE/CVF Winter Conference on Applications of Computer Vision (WACV), IEEE. doi:10.1109/WACV57701.2024.00799.

Rajah, P., Odindi, J., Mutanga, O., and Kiala, Z. 2019. “The utility of Sentinel-2 Vegetation Indices (VIs) and Sentinel-1 Synthetic Aperture Radar (SAR) for invasive alien species detection and mapping.” *Nature Conservation*, Vol. 35, pp. 41–61. doi:10.3897/natureconservation.35.29588.

Ramsar Sites Information Service [WWW Document]. 2001. https://rsis.ramsar.org/ris/949 (accessed 12.16.25).

Rand, W.M. n.d. “Objective Criteria for the Evaluation of Clustering Methods.”

Rousseeuw, P.J. 1987. “Silhouettes: A graphical aid to the interpretation and validation of cluster analysis.” *Journal of Computational and Applied Mathematics*, Vol. 20, pp. 53–65. doi:10.1016/0377-0427(87)90125-7.

Settles, B. n.d. *Active Learning Literature Survey* no. 1648. University of Wisconsin–Madison.

Sierra, S., Ramo, R., Padilla, M., Quirós, L., and Cobo, A. 2025. “Estimating Fractional Land Cover Using Sentinel-2 and Multi-Source Data with Traditional Machine Learning and Deep Learning Approaches.” *Remote Sensing*, Vol. 17 (No. 19), p. 3364. doi:10.3390/rs17193364.

Tsendbazar, N. 2022. *Product Validation Report (D12-PVR)* nos. D12-PVR. ESA WorldCover.

Vaswani, A., Shazeer, N., Parmar, N., Uszkoreit, J., Jones, L., Gomez, A.N., Kaiser, L., and Polosukhin, I. 2023. “Attention Is All You Need.” doi:10.48550/arXiv.1706.03762.

Venter, Z.S., Barton, D.N., Chakraborty, T., Simensen, T., and Singh, G. 2022. “Global 10 m Land Use Land Cover Datasets: A Comparison of Dynamic World, World Cover and Esri Land Cover.” *Remote Sensing*, Vol. 14 (No. 16), p. 4101. doi:10.3390/rs14164101.

Xie, E., Wang, W., Yu, Z., Anandkumar, A., Alvarez, J.M., and Luo, P. 2021. “SegFormer: Simple and Efficient Design for Semantic Segmentation with Transformers.” doi:10.48550/arXiv.2105.15203.

Xu, P., Tsendbazar, N.-E., Herold, M., De Bruin, S., Koopmans, M., Birch, T., Carter, S., Fritz, S., Lesiv, M., Mazur, E., Pickens, A., Potapov, P., Stolle, F., Tyukavina, A., Van De Kerchove, R., and Zanaga, D. 2024. “Comparative validation of recent 10 m-resolution global land cover maps.” *Remote Sensing of Environment*, Vol. 311, p. 114316. doi:10.1016/j.rse.2024.114316.

Yi, Y., Zhang, Z., Zhang, W., Zhang, C., Li, W., and Zhao, T. 2019. “Semantic Segmentation of Urban Buildings from VHR Remote Sensing Imagery Using a Deep Convolutional Neural Network.” *Remote Sensing*, Vol. 11 (No. 15), p. 1774. doi:10.3390/rs11151774.

Zanaga, D., Van De Kerchove, R., Daems, D., De Keersmaecker, W., Brockmann, C., Kirches, G., Wevers, J., Cartus, O., Santoro, M., Fritz, S., Lesiv, M., Herold, M., Tsendbazar, N.-E., Xu, P., Ramoino, F., and Arino, O. 2022. “ESA WorldCover 10 m 2021 v200.” doi:10.5281/ZENODO.7254221.

Zhao, J., Wang, L., Yang, H., Wu, P., Wang, B., Pan, C., and Wu, Y. 2022. “A Land Cover Classification Method for High-Resolution Remote Sensing Images Based on NDVI Deep Learning Fusion Network.” *Remote Sensing*, Vol. 14 (No. 21), p. 5455. doi:10.3390/rs14215455.

Zhao, S., Tu, K., Ye, S., Tang, H., Hu, Y., and Xie, C. 2023. “Land Use and Land Cover Classification Meets Deep Learning: A Review.” *Sensors*, Vol. 23 (No. 21), p. 8966. doi:10.3390/s23218966.

Zhu, X.X., Tuia, D., Mou, L., Xia, G.-S., Zhang, L., Xu, F., and Fraundorfer, F. 2017. “Deep Learning in Remote Sensing: A Comprehensive Review and List of Resources.” *IEEE Geoscience and Remote Sensing Magazine*, Vol. 5 (No. 4), pp. 8–36. doi:10.1109/MGRS.2017.2762307.

## Data Availability

The COTQ land-cover map and the reference products used in this study are publicly available for visual inspection at https://ee-dga-couverture.projects.earthengine.app/view/cotq. Sentinel-2 imagery is distributed by the European Space Agency. Derived evaluation metrics and scripts are available from the authors upon reasonable request.

## Declaration of Interest

The authors declare that they have no known competing financial or non-financial interests that could have appeared to influence the work reported in this paper. The Ministère des Ressources naturelles et des Forêts, which supported the development of the COTQ product, had no involvement in the study design, data analysis, interpretation of the results, or in the writing of the manuscript.

## Use of Generative Artificial Intelligence

Generative artificial intelligence tools were used in the preparation of this manuscript. Specifically, AI-based language models were employed to assist with the translation of the manuscript from French to English and to improve linguistic clarity. In addition, generative AI was partially used to support the development of certain code components utilized in the analyses. All outputs generated with the assistance of AI were carefully reviewed, verified, and edited by the authors, who take full responsibility for the content, accuracy, and integrity of the manuscript and associated code.

## Funding

This work was supported by the Ministère des Ressources Naturelles et des Forêts under Grant 610024-04.

## Acknowledgements

The authors acknowledge the support of the Ministère des Ressources naturelles et des Forêts (MRNF) in the context of the COTQ project, which provided the institutional framework for this study.

## Author contributions statement

Conceptualization: E.C., S.F., Y.B.

Methodology: E.C.

Software: E.C.

Analysis: E.C.

Writing - original draft: E.C.

Writing - review & editing: E.C., S.F., Y.B.